\documentclass[manuscript,screen]{acmart}
\AtBeginDocument{%
  }
\makeatletter\let\showhyphens\@undefined\makeatother
\usepackage[ruled,vlined,linesnumbered]{algorithm2e}
\newtheorem{definition}{Definition}
\setcopyright{cc}
\copyrightyear{2026}
\acmYear{2026}
\acmDOI{XXXXXXX.XXXXXXX}
\acmJournal{JRC}
\acmVolume{0}
\acmNumber{0}
\acmArticle{0}
\acmMonth{0}
\newcommand{\toolname}{LLM-FACETS}
\begin{document}
\title[LLM-FACETS: Privacy-Preserving LLM Evaluation Framework]{LLM-FACETS: A Privacy-Preserving Framework for Evaluating LLM Transparency and Accountability}
\titlenote{Source code available at \url{https://github.com/Scriptor-Group/AIMVi}.}
\author{Tom Lucas}
\authornote{Corresponding author.}
\email{tom@devana.ai}
\orcid{0009-0000-8357-3477}
\affiliation{\institution{Luxembourg Institute of Science and Technology (LIST)}\city{Esch-sur-Alzette}\country{Luxembourg}}
\affiliation{\institution{University of Luxembourg}\city{Esch-sur-Alzette}\country{Luxembourg}}
\affiliation{\institution{Scriptor Artis}\city{Maz\`{e}res}\country{France}}
\author{Alessio Buscemi}
\email{alessio.buscemi@list.lu}
\orcid{0000-0001-9765-1907}
\affiliation{\institution{Luxembourg Institute of Science and Technology (LIST)}\city{Esch-sur-Alzette}\country{Luxembourg}}
\author{Alfredo Capozucca}
\email{alfredo.capozucca@uni.lu}
\orcid{0009-0003-4668-9915}
\affiliation{\institution{University of Luxembourg}\city{Esch-sur-Alzette}\country{Luxembourg}}
\author{German Castignani}
\email{german.castignani@list.lu}
\orcid{0000-0001-5594-4904}
\affiliation{\institution{Luxembourg Institute of Science and Technology (LIST)}\city{Esch-sur-Alzette}\country{Luxembourg}}
\author{Barbara Delacroix}
\email{barbara@scriptorartis.com}
\affiliation{\institution{Scriptor Artis}\city{Maz\`{e}res}\country{France}}
\renewcommand{\shortauthors}{Lucas et al.}
\begin{abstract}
Assessing whether Large Language Models outputs are factually grounded, epistemically calibrated, and methodologically reproducible is a prerequisite for responsible AI deployment. Yet the practice of auditing LLMs remains difficult to access for non-technical practitioners: existing tools require programming expertise, non-trivial environment configuration, and lack systematic methodological approaches to multi-dimensional evaluation, while cloud-hosted platforms transmit evaluation data to external services, creating barriers for domain experts and compliance officers who bear legal and ethical responsibility for AI oversight. We introduce \textit{LLM-FACETS} (LLM \textbf{F}Actuality \textbf{C}ross-Evalua\textbf{T}ion \textbf{S}ystem): an open-source evaluation framework, accompanied by a tool offering a browser-accessible interface and a plugin architecture, designed to help practitioners make informed, transparent decisions about LLM quality. The framework structures evaluation around three practitioner profiles (technical experts, domain experts, and compliance officers), inspired by the stakeholder categories identified in the EU AI Act and the NIST AI Risk Management Framework as responsible for human oversight of AI systems. The tool's architecture makes data flows explicit: deterministic metrics (BLEU, ROUGE, BERTScore) run entirely within the self-hosted server process with no outbound data transmission; LLM-judge metrics contact external APIs explicitly, with users retaining full control over credentials and responsible for applicable data protection agreements. The framework operationalizes transparency through three auditing mechanisms: token-level log-probability visualization for epistemic uncertainty (model confidence) assessment, multi-judge consensus evaluation to mitigate judge bias, and RAG Triad metrics (Faithfulness, Answer Relevance, Context Relevance) to detect and localize hallucinations. A plugin architecture allows any new metric or dataset to be integrated without modifying the evaluation pipeline. The open-source implementation enables cross-checking across multiple metrics targeting the same property, ensuring the reproducibility required by open science principles, and decoupling AI accountability from the teams that built the systems being assessed. We verify the framework through cross-validation of 18~metric implementations against canonical reference libraries.
\end{abstract}
\maketitle
\section{Introduction}
Auditing Large Language Models (LLMs)---assessing whether their outputs are factually grounded, epistemically calibrated, and methodologically reproducible---has become a prerequisite for responsible deployment across high-stakes domains, from clinical decision support to legal document analysis~\cite{euaiact2024}. Yet this practice remains largely inaccessible to the practitioners who need it most. The tools required demand programming expertise, non-trivial environment configuration, and data transmission to external APIs. As a result, the actors most responsible for accountability, e.g. domain experts and compliance officers, are often excluded from evaluation processes they are legally and ethically required to perform.
Regulatory frameworks make this gap difficult to ignore. The EU Artificial Intelligence Act (AI Act)~\cite{euaiact2024} requires that human overseers of high-risk AI systems be capable of interpreting model outputs and certifying compliance, which is, in practice, a functional requirement for evaluation interfaces that should be accessible also for non-engineers. The General Data Protection Regulation (GDPR)~\cite{gdpr2016} requires a formal Data Processing Agreement whenever personal data is transmitted to a third-party processor, raising compliance questions for cloud-hosted evaluation platforms applied to sensitive datasets.
Existing evaluation ecosystems reinforce this fragmentation. Programmatic evaluation libraries such as DeepEval~\cite{deepeval2024} and Ragas~\cite{ragas2024} offer powerful evaluation but require programming expertise, dependency management, and scripting---Barrier~1: programming barrier; Barrier~2: environment configuration. Together, these two barriers result in the systematic exclusion of non-technical practitioners from evaluation workflows. On the hosted-service side, platforms such as Arize Phoenix~\cite{arizephoenix2024} and Langfuse~\cite{langfuse2024} focus on monitoring deployed LLM applications in live production environments rather than collaborative metric exploration. Academic visual analytics tools~\cite{kahng2024} provide sophisticated visualizations but remain isolated prototypes. None of these solutions addresses data sovereignty: evaluating sensitive datasets (medical records, legal documents, proprietary corpora) through third-party APIs creates irreconcilable conflicts with data protection regulations---Lack of data sovereignty assurance.
This paper presents \textit{LLM-FACETS}, an open-source evaluation framework that directly addresses these barriers. The framework is structured around three practitioner profiles (technical experts, domain experts, and compliance officers). This mirrors the functional roles identified in the EU AI Act Article~14~\cite{euaiact2024} and the NIST AI Risk Management Framework~\cite{nistairmf2023} as responsible for human oversight of AI systems. The framework makes three primary contributions:
\begin{enumerate}
  \item A \textbf{methodological framework} connecting transparency goals to concrete evaluation practices, structured around the aforementioned practitioner profiles. Each profile has distinct transparency needs, accountability roles, and recommended metric configurations.
  \item A \textbf{unified suite of transparency auditing mechanisms} spanning epistemic uncertainty (log-probability visualization), factual grounding (RAG Triad), and process fairness (multi-judge consensus), delivered through a browser-accessible interface that requires no programming expertise.
  \item An \textbf{open-source tool} implementing the framework through 18 metric variants in TypeScript, cross-validated against canonical Python reference implementations, with a plugin architecture in which adding a new metric or dataset propagates automatically to the navigation, benchmarking dashboard, and REST API without further configuration\footnote{Source code: \url{https://github.com/Scriptor-Group/AIMVi}.}---enabling reproducible and auditable evaluation independent of proprietary tooling (Section~\ref{sec:verification}).
\end{enumerate}
The remainder of this paper is structured as follows. Section~\ref{sec:challenges} identifies the challenges inherent to LLM evaluation, including the need for transparency across multiple dimensions. Section~\ref{sec:audience} defines the three practitioner profiles, formalizes the framework, and presents the evaluation workflow. Section~\ref{sec:architecture} describes the LLM-FACETS tool architecture. Section~\ref{sec:metrics} details how the implemented metrics operationalize the transparency dimensions. Section~\ref{sec:verificationperf} provides cross-validation evidence and performance analysis. Section~\ref{sec:usecases} validates the proposed framework through multi-stakeholder use cases. Section~\ref{sec:discussion} discusses limitations and broader impact. Section~\ref{sec:conclusion} concludes.
\section{Challenges}
\label{sec:challenges}
This section identifies the five challenges that motivate this work: the fragmented landscape of existing evaluation tools~(\S\ref{sec:c1}), the design requirements for accessible AI explanation interfaces~(\S\ref{sec:c2}), the legal and regulatory constraints on evaluation practices~(\S\ref{sec:c3}), the reproducibility crisis in LLM benchmarking~(\S\ref{sec:c4}), and the multi-dimensional nature of transparency itself~(\S\ref{sec:transparencydimensions}). Together, these challenges establish the concrete requirements that the framework must satisfy.
\subsection{C1 --- Onboarding}
\label{sec:c1}
The rapid adoption of LLMs has spurred a proliferation of evaluation tools, each addressing a subset of the evaluation challenge---broadly defined as the problem of systematically measuring model output quality in a reproducible, interpretable, and scalable manner~\cite{gehrmann2022repairing}. Frameworks addressing factual accuracy and targeted evaluation exist. For instance, Ragas~\cite{ragas2024} provides a programmatic framework for evaluating Retrieval Augmented Generation (RAG) pipelines through metrics such as Faithfulness, Answer Relevance, and Context Relevance. DeepEval~\cite{deepeval2024} extends this to a broader suite of over 14 metrics including hallucination detection~\cite{li2023halueval}, bias assessment~\cite{zheng2023judging}, and toxicity scoring, each grounded in peer-reviewed evaluation research. Both frameworks are programmatic: they require Python environments, dependency management, and scripting to execute evaluations.
On the hosted-service side, LangSmith~\cite{langsmith2023} offers a cloud-hosted observability platform focused on tracing and monitoring deployed LLM applications in live production environments, offering custom LLM-as-a-Judge templates rather than a standardized metric suite. TruLens~\cite{trulens2024} provides feedback functions for evaluating RAG applications with a visualization dashboard. Arize Phoenix and Langfuse operate in a similar space, focusing on production monitoring and trace analysis.
These tools share two structural limitations. First, they present a \textit{technical barrier}: all require programming expertise and development environment configuration, making participation difficult for domain experts and compliance officers. Second, cloud-hosted platforms such as LangSmith transmit evaluation data---which may include sensitive documents---to external servers, creating data sovereignty questions. Even nominally open-source tools that rely on cloud-based LLM judge calls transmit the texts being evaluated to third-party providers. This dual limitation---technical complexity and data exposure---motivates the design of a tool that is both browser-accessible and explicit about where data flows.
A complementary approach is taken by ARES~\cite{ares2024}, which trains lightweight discriminative judges (DeBERTa variants) specifically for RAG evaluation tasks. By fine-tuning small models on domain-relevant data, ARES reduces dependence on frontier model APIs while achieving comparable precision at a fraction of the cost---a significant advantage for high-volume or cost-constrained evaluation. However, ARES remains a Python research framework without a visual interface or data sovereignty guarantees.
Academic visual analytics tools have begun to address the accessibility dimension. LLM Comparator~\cite{kahng2024} provides side-by-side visual evaluation of LLM outputs for qualitative comparison. However, such tools remain isolated prototypes that do not integrate quantitative metrics, privacy guarantees, or multi-practitioner workflows into a unified evaluation framework. The lack of such unified tools represents a major onboarding and accessibility barrier for institutions.
\subsection{C2 --- Accessibility}
\label{sec:c2}
A growing body of HCI research demonstrates that the design of AI explanation interfaces critically determines whether practitioners can effectively exercise oversight. Liao et al.~\cite{liao2020questioning} establish that AI explanation tools must be designed around the questions that practitioners actually ask---not around the technical artifacts that engineers find convenient to produce. Their question-driven design framework reveals a persistent mismatch: most explanation tools answer ``how does the model work?'' when users need to know ``can I trust this specific output?''
Ehsan et al.~\cite{ehsan2021expanding} extend this insight to the concept of \textit{social transparency}: making the reasoning processes and social contexts behind AI decisions accessible to non-technical practitioners. Their work demonstrates that transparency is not achieved by exposing model internals, but by presenting evaluation processes in forms that enable meaningful participation by diverse actors.
For LLM evaluation tooling, the implication is concrete: presenting results as raw numerical scores in terminal outputs or Jupyter notebooks fails the accessibility test that this body of research identifies as essential for genuine human oversight~\cite{liao2020questioning,ehsan2021expanding}. An evaluation framework that aims to support multi-practitioner auditing must therefore translate scores into interactive representations structured around the questions that non-technical auditors actually ask: ``Is this output factually grounded?'', ``Can I trust this specific claim?'', ``Is the evaluation methodology itself reliable?''~\cite{liao2020questioning}. Producing tools capable of empowering domain experts without sacrificing technical depth is a critical challenge.
\subsection{C3 --- Compliance}
\label{sec:c3}
The global regulatory landscape increasingly mandates evaluation capabilities that existing tools do not provide. While these requirements take different forms across jurisdictions (such as the proposed US Algorithmic Accountability Act), the European legal framework provides explicit, broadly applicable examples of these duties. The EU AI Act~\cite{euaiact2024} imposes multiple evaluation-relevant obligations on providers and deployers of high-risk AI systems. Articles~9 and 11 require continuous risk assessment and detailed technical documentation of evaluation metrics and their results. Article~13 mandates that system outputs be interpretable by their intended users. Article~14 requires human oversight by persons capable of understanding the capabilities and limitations of the system---a functional requirement for accessible evaluation interfaces.
The GDPR~\cite{gdpr2016} introduces a distinct but equally binding constraint. Article~28 requires a formal Data Processing Agreement (DPA) whenever personal data is transmitted to a third party for processing. Evaluating LLM outputs on datasets containing personal information---patient records, employee communications, legal case files---through cloud-hosted evaluation APIs constitutes data processing under this definition. In the absence of a DPA with the evaluation platform provider, such evaluation is unlawful under European data protection law.
ISO/IEC~42001:2023~\cite{iso42001} provides the first international management system standard for AI, requiring organizations to establish documented processes for AI risk assessment, continuous monitoring, and human control. The NIST AI Risk Management Framework~\cite{nistairmf2023} similarly recommends measurable evaluation practices accessible to diverse practitioners. Taken together, these three instruments establish distinct but complementary requirements: the EU AI Act mandates accessibility (Art.~14) and technical documentation (Arts.~9, 11, 13); GDPR constrains data transmission to third parties (Art.~28); ISO~42001 requires documented, continuous monitoring processes. No existing evaluation tool satisfies all three requirements simultaneously---a gap that motivates the architectural choices described in Section~\ref{sec:architecture}.
Commercial governance platforms such as Credo~AI~\cite{credoai2024} and Fiddler~AI~\cite{fiddlerai2024} address the compliance reporting dimension by translating technical evaluation evidence into regulatory artifacts mapped to the EU AI Act and ISO~42001. However, these platforms require cloud data transmission and do not address the data sovereignty constraint: organizations subject to GDPR cannot use them to evaluate datasets containing personal information without a Data Processing Agreement. This leaves a structural gap between compliance tooling and privacy-preserving evaluation.
\subsection{C4 --- Reproducibility}
\label{sec:c4}
The reliability of LLM evaluation itself is increasingly questioned. Gehrmann et al.~\cite{gehrmann2022repairing} conduct a comprehensive survey documenting evaluation obstacles including inconsistent metric implementations, unreported hyperparameters, and lack of standardization---concluding that the evaluation foundation of NLP (Natural Language Processing) exhibits systemic vulnerabilities that require methodological repair. Post~\cite{post2018call} demonstrates that reported BLEU scores---a widely used lexical overlap metric measuring $n$-gram precision between a generated text and reference translations---are not comparable across publications because implementations differ in tokenization, smoothing, and case normalization, rendering cross-paper comparisons unreliable.
Questioning the evaluation paradigm itself, Bender et al.~\cite{bender2021dangers} argue that the benchmarking paradigm creates a false sense of progress: language model outputs are optimized for benchmark performance without adequate evaluation of real-world behavior, and the scale of modern language models makes meaningful evaluation increasingly difficult. This critique gains urgency in the LLM-as-a-Judge paradigm, where Zheng et al.~\cite{zheng2023judging} quantify position bias at 5--15~percentage points and demonstrate that single-judge evaluation produces unstable rankings.
Three requirements for credible LLM evaluation follow from these findings:
\begin{enumerate}
  \item Deterministic metric implementations must be cross-validated to ensure comparability across tools and studies.
  \item LLM judge evaluations must report variance and bias quantification alongside primary scores.
  \item Evaluation methodology must be transparent and reproducible so that third parties can verify published results.
\end{enumerate}
Building tools that enforce these requirements systematically, rather than leaving them to ad-hoc scripts, is essential to repair the evaluation foundation.
\subsection{C5 --- Transparency}
\label{sec:transparencydimensions}
Transparency in AI is a multidimensional concept that is frequently conflated with a single technical property (e.g., explainability or interpretability). A more fundamental observation, however, is that transparency is not a static property that a system either has or lacks: it is an active \textit{relation} between information and its audience~\cite{liao2020questioning,ehsan2021expanding}. The same model output that is transparent to a machine learning engineer reading raw log-probabilities in a terminal is entirely opaque to a compliance officer who needs to certify whether the model fabricates medical claims. Transparency, in other words, is always transparency \textit{for someone}, shaped by that person's epistemic needs, accountability role, and available tools.
This relational framing has a direct architectural consequence: a framework that genuinely engineers transparency cannot offer a single view of model output quality and reliability. It must provide distinct mechanisms oriented toward distinct purposes---each designed around the questions that a specific audience needs to answer. For the purposes of LLM evaluation, we distinguish four operationalizable dimensions that correspond directly to the framework's feature set.
\subsubsection{Epistemic Transparency: Knowing What the Model Doesn't Know}
Epistemic transparency concerns the model's ability to communicate its own uncertainty. A model that confidently generates a plausible-sounding but factually incorrect statement---a hallucination---fails epistemically not because it errs, but because it does not signal that it may err. Token-level log-probabilities (LogProbs) provide a direct window into this dimension. A \textit{token} is a sub-word unit (e.g., a word or morpheme) produced sequentially by the model's decoding process. For each generated token $t_i$, the model assigns an output probability $p(t_i \mid t_{<i})$: the conditional probability of that token given all preceding tokens. Low values indicate regions of uncertainty where hallucination risk is elevated. The sequence-level confidence can be summarized as the mean of per-token log-probabilities: $\frac{1}{n}\sum_{i=1}^{n} \log p(t_i \mid t_{<i})$.
Prior academic work has addressed LogProb visualization for ML researchers: the LM Transparency Tool ~\cite{lmtt2024} provides interactive projections of layer-wise attention contributions and token probability distributions, enabling mechanistic analysis of Transformer internals. However, such tools target researchers studying model behavior, not non-technical auditors assessing the reliability of specific outputs in a production context.
Despite their diagnostic value, LogProbs remain underutilized in practice because, to our knowledge, no standard web interface exposes them in an accessible, visual format for non-technical practitioners. Creating such interfaces is a critical challenge.
\subsubsection{Factual Transparency: Tracing Outputs to Sources}
Factual transparency is central to RAG systems, which are increasingly deployed to ground LLM outputs in retrieved documents. When such a system generates a claim, a responsible evaluator must ask: \textit{Is this claim supported by the retrieved context? Is the retrieved context actually relevant to the question? Is the generated answer actually addressing the question asked?}
The RAG Triad---Faithfulness, Answer Relevance, and Context Relevance~\cite{ragas2024}---operationalizes these three questions as measurable scores. Faithfulness decomposes the generated response into atomic claims and verifies each against the retrieved context. Answer Relevance uses reverse question generation to assess whether the response actually addresses the original query. Context Relevance classifies retrieved sentences as signal or noise relative to the query. Together, these metrics provide a traceable audit trail linking the model output back to its evidential basis.
\subsubsection{Process Transparency: Auditing the Evaluators}
A less-discussed but critical form of transparency concerns the evaluation process itself. LLM-as-a-Judge methods---where a powerful LLM scores the outputs of another model---have become the dominant paradigm for automated quality assessment~\cite{zheng2023judging,liu2023geval}. However, these judges are subject to well-documented biases: position bias (favoring responses presented first in the prompt context, meaning the comparison $A$ vs $B$ may yield a different outcome than $B$ vs $A$ solely due to their spatial ordering in the LLM's context window), verbosity bias (rewarding longer responses), and self-enhancement bias (favoring outputs from the same model family).
Process transparency requires that the evaluation methodology itself be auditable. To enable practitioners to assess not just \textit{what} the evaluation result is, but \textit{how reliable the evaluation process itself is}, tools must mitigate known biases like position-bias and verbosity bias across comparisons, and ensure the variance of results is exposed. Without mechanisms like multi-judge consensus or prompt permutations transparently exposed, automated evaluation remains a black box for non-technical practitioners.
\subsubsection{Baseline Grounding: Anchoring Evaluation in Deterministic References}
A fourth dimension of transparency concerns the availability of deterministic, reproducible reference points against which LLM-based evaluations can be calibrated. Traditional n-gram metrics (BLEU, ROUGE, METEOR) and neural similarity measures (BERTScore) do not themselves implement epistemic, factual, or process transparency in the senses defined above. Instead, they provide \textit{baseline grounding}: well-understood, externally validated measurements whose outputs are fully determined by the input---without stochastic LLM calls---and whose implementations can be cross-validated against canonical reference libraries. By anchoring the evaluation in these deterministic baselines, auditors can detect systematic divergences between traditional scores and LLM judge scores, which may reveal prompt engineering artifacts, model self-bias, or genuine quality differentials that word-overlap metrics cannot capture. Baseline grounding thus serves as a calibration layer that strengthens the credibility of the other three transparency dimensions.
\section{Audience, terminology, and workflow}
\label{sec:audience}
The value of an evaluation framework is realized only when its methods are genuinely usable by the practitioners who need them. This section defines the methodological contribution that structures how practitioners apply evaluation features purposefully to produce accountable evidence. It first characterizes the three practitioner profiles, then formalizes the framework as a mathematical structure, and finally presents the evaluation workflow.
\subsection{Audience}
\label{sec:audience-profiles}
To address the diverse accountability demands of LLM evaluation, the framework is built around three practitioner profiles. These profiles map specific responsibilities to actionable transparency goals:
\begin{table}[h]
  \caption{Practitioner Profiles and their corresponding transparency needs}
  \label{tab:profiles}
  \small
  \begin{tabular}{p{2.5cm} p{3.5cm} p{2.5cm} p{3cm} p{3cm}}
    \toprule
    \textbf{Profile} & \textbf{Goal} & \textbf{Transparency Needs} & \textbf{Recommended Metrics} & \textbf{Expected Output} \\
    \midrule
    Technical Expert & Diagnose hallucination origins & Epistemic, Factual & RAG Triad, LogProbs & Per-claim breakdown \\
    Domain Expert & Assess reliability within domain & Factual, Process & Faithfulness, Jury & Structured reports \\
    Compliance Officer & Regulatory certification & Process & RAG Triad, Batch Faithfulness & JSON/CSV audit trails \\
    \bottomrule
  \end{tabular}
\end{table}
The selection of these three profiles is grounded in the functional responsibilities that regulatory instruments assign to distinct categories of human overseers. Article~14 of the EU AI Act~\cite{euaiact2024} requires that persons exercising oversight of high-risk AI systems possess the ability to ``correctly interpret the high-risk AI system's output,'' a competence that differs structurally depending on whether the overseer is a technical implementer, a subject-matter authority, or a governance officer. The NIST AI Risk Management Framework~\cite{nistairmf2023} similarly distinguishes between AI actors who \textit{design and develop} AI systems, those who \textit{deploy and use} them within a domain of expertise, and those responsible for \textit{governance and compliance}---three functional categories that map directly onto the profiles in Table~\ref{tab:profiles}.
Table~\ref{tab:profiles} summarizes how each profile's accountability role determines a distinct transparency need and, consequently, a distinct subset of the framework's metrics. The \textit{Technical Expert} (e.g., an ML engineer debugging a RAG pipeline) requires epistemic and factual transparency: token-level log-probabilities to diagnose model uncertainty, and the RAG Triad to trace hallucinations to retrieval failures, producing per-claim diagnostic breakdowns. The \textit{Domain Expert} (e.g., a clinical pharmacologist validating drug-interaction answers) requires factual and process transparency: Faithfulness scores to verify that outputs are grounded in authoritative sources, and the Jury module to assess whether evaluation judgments are stable across multiple independent judges, producing structured reports suitable for communication with technical teams. The \textit{Compliance Officer} (e.g., a data protection officer certifying a patient triage system) requires process transparency above all: batch-level RAG Triad scores and Faithfulness audits that can be exported as JSON/CSV audit trails for inclusion in regulatory documentation required under Articles~9 and~11 of the EU AI Act.
These profiles are not mutually exclusive: a single evaluation session may serve multiple roles simultaneously. They are, however, distinct in the \textit{purpose} they assign to transparency---diagnostic, deliberative, or compliance-oriented---and the framework's design ensures that each purpose maps to specific, usable evaluation mechanisms.
\subsection{Terminology and Framework Definition}
\label{sec:frameworkdef}
\begin{definition}
The \emph{LLM-FACETS} framework is a four-tuple
\[
  \mathcal{F} = (P,\; M,\; W,\; E)
\]
where:
\begin{itemize}
  \item $P = \{p_{\mathrm{eng}},\, p_{\mathrm{dom}},\, p_{\mathrm{cmp}}\}$ is the finite set of \emph{practitioner profiles}---technical expert, domain expert, and compliance officer---each characterized by a transparency goal $g(p) \in \{\mathrm{baseline\_grounding},\, \mathrm{epistemic},\, \mathrm{factual},\, \mathrm{process}\}$, a recommended metric subset $M_p \subseteq M$, and an expected output form $o(p)$;
  \item $M = M_{\mathrm{lex}} \cup M_{\mathrm{neu}} \cup M_{\mathrm{judge}} \cup M_{\mathrm{rag}}$ is the \emph{set of metrics} (Table~\ref{tab:metrics}), where each metric $m \in M$ is annotated with a transparency dimension $d(m) \in \{\mathrm{baseline\_grounding},\, \mathrm{epistemic},\, \mathrm{factual},\, \mathrm{process}\}$ and an execution type (deterministic, provider-native, or LLM-based);
  \item $W = (w_1, \ldots, w_5)$ is the \emph{workflow} (presented in the next subsection), mapping a profile $p \in P$ and a dataset $D$ to a scored evidence set $\mathcal{E}$;
  \item $E$ is the \emph{generated evidence}, producing structured artifacts containing per-sample scores, metric configurations, and full reasoning chains, enabling independent verification by third parties.
\end{itemize}
\end{definition}
The four components interact as follows. A practitioner selects a profile $p \in P$. The workflow $W$ guides the practitioner through five ordered steps---from goal definition to evidence export---enforcing the cross-checking property that at least two metrics targeting the same dimension are used. The audit interface $E$ packages the resulting evidence into a portable, independently verifiable artifact.
\subsection{Evaluation Workflow}
\label{sec:workflow}
The evaluation workflow is defined independently of any specific tool implementation. It structures how any practitioner---regardless of technical background---should proceed to obtain accountable evidence that allows them to make informative decisions:
\begin{enumerate}
  \item \textbf{Define the transparency goal} (Algorithm~\ref{alg:workflow}, Lines~1--11). Identify which dimension is at stake---epistemic (model confidence), factual (output grounding), or process (evaluation reliability)---using the taxonomy in Section~\ref{sec:transparencydimensions} as a guide.
  \item \textbf{Select relevant metrics} (Algorithm~\ref{alg:workflow}, Lines~12--14). Consult Table~\ref{tab:metrics} to identify the metrics that operationalize the target dimension. For robust conclusions, select at least two metrics targeting the same property to enable cross-checking.
  \item \textbf{Load or configure a dataset} (Algorithm~\ref{alg:workflow}, Lines~15--16). Use an integrated benchmark (SQuAD~v2, SelfAware, HaluEval) for standardized comparability, or upload a proprietary dataset. The architecture (Section~\ref{sec:architecture}) ensures that deterministic metrics require no outbound data transmission regardless of dataset sensitivity.
  \item \textbf{Cross-check results} (Algorithm~\ref{alg:workflow}, Lines~17--25). Interpret scores from multiple metrics targeting the same property. Convergence provides stronger evidence; divergence reveals where metrics disagree and warrants deeper investigation. When metrics targeting the same dimension diverge ($r < 0.5$ in the correlation heatmap), examine intermediate outputs: for Faithfulness vs.\ BERTScore divergence, inspect the claim-decomposition reasoning chain for errors. The benchmark analysis dashboard (Section~\ref{sec:architecture}) provides correlation heatmaps and distribution plots for this purpose.
  \item \textbf{Package evidence} (Algorithm~\ref{alg:workflow}, Lines~26--27). When evaluation is complete, results can be exported as structured JSON or CSV as a portable audit trail. The exported file contains per-sample scores, metric configurations, and---for LLM-as-a-Judge metrics---the full reasoning chains, enabling independent verification by third parties.
\end{enumerate}
Algorithm~\ref{alg:workflow} formalizes this workflow, making explicit the inputs, outputs, and decision points of each step.
\begin{algorithm}[t]
\caption{LLM-FACETS Evaluation Workflow}
\label{alg:workflow}
\DontPrintSemicolon
\SetAlgoLined
\KwIn{Practitioner profile $p \in P$;\; dataset $D$ (benchmark or proprietary corpus)}
\KwOut{Evidence set $\mathcal{E}$: per-sample scores, configurations, and reasoning chains}
\BlankLine
\textbf{Step 1 --- Choose an LLM}\;
\Indp
$l \leftarrow \textsc{SelectModel}()$\tcp*{Select provider and model}
\Indm
\BlankLine
\textbf{Step 2 --- Choose if anonymisation or not}\;
\Indp
$a \leftarrow \textsc{SelectAnonymisation}()$\tcp*{Local NER / Presidio / None}
\Indm
\BlankLine
\textbf{Step 3 --- Define transparency goal}\;
\Indp
$d \leftarrow g(p)$\tcp*{epistemic / factual / process (Section~\ref{sec:transparencydimensions})}
\Indm
\BlankLine
\textbf{Step 4 --- Select metric subset}\;
\Indp
$M_p \leftarrow \{m \in M \mid d(m) = d\}$\;
\lIf{$|M_p| < 2$}{\textbf{warn}: cross-checking requires ${\geq}\,2$ metrics per dimension}
\Indm
\BlankLine
\textbf{Step 5 --- Load and configure dataset}\;
\Indp
\lIf{$D$ is a benchmark}{load via Dataset Registry (Section~\ref{sec:architecture})}
\lElse{upload CSV/Parquet; query in-process via DuckDB (data sovereignty preserved)}
\lIf{$D$ contains Personally Identifiable Information (PII)}{apply anonymization pipeline before any outbound API call}
\Indm
\BlankLine
\textbf{Step 6 --- Cross-check results}\;
\Indp
\ForEach{$m \in M_p$, $s \in D$}{
  $\mathrm{score}_{m}(s) \leftarrow \textsc{Evaluate}(m, s)$\;
}
Compute inter-metric Pearson correlation matrix $R$ over $\{\mathrm{score}_{m}\}_{m \in M_p}$\;
\lIf{$\exists\,m_i, m_j \in M_p :\ r_{ij} < 0.5$}{flag divergence; inspect intermediate outputs}
\Indm
\BlankLine
\textbf{Step 7 --- Package evidence (optional)}\;
\Indp
$\mathcal{E} \leftarrow \textsc{Export}(\{\mathrm{score}_{m}\}_{m \in M_p},\; \text{configurations},\; \text{reasoning chains})$\tcp*{JSON / CSV}
\Return $\mathcal{E}$
\Indm
\end{algorithm}
\section{LLM-FACETS Tool}
\label{sec:architecture}
\begin{figure}[h]
\centering
\includegraphics[width=\linewidth]{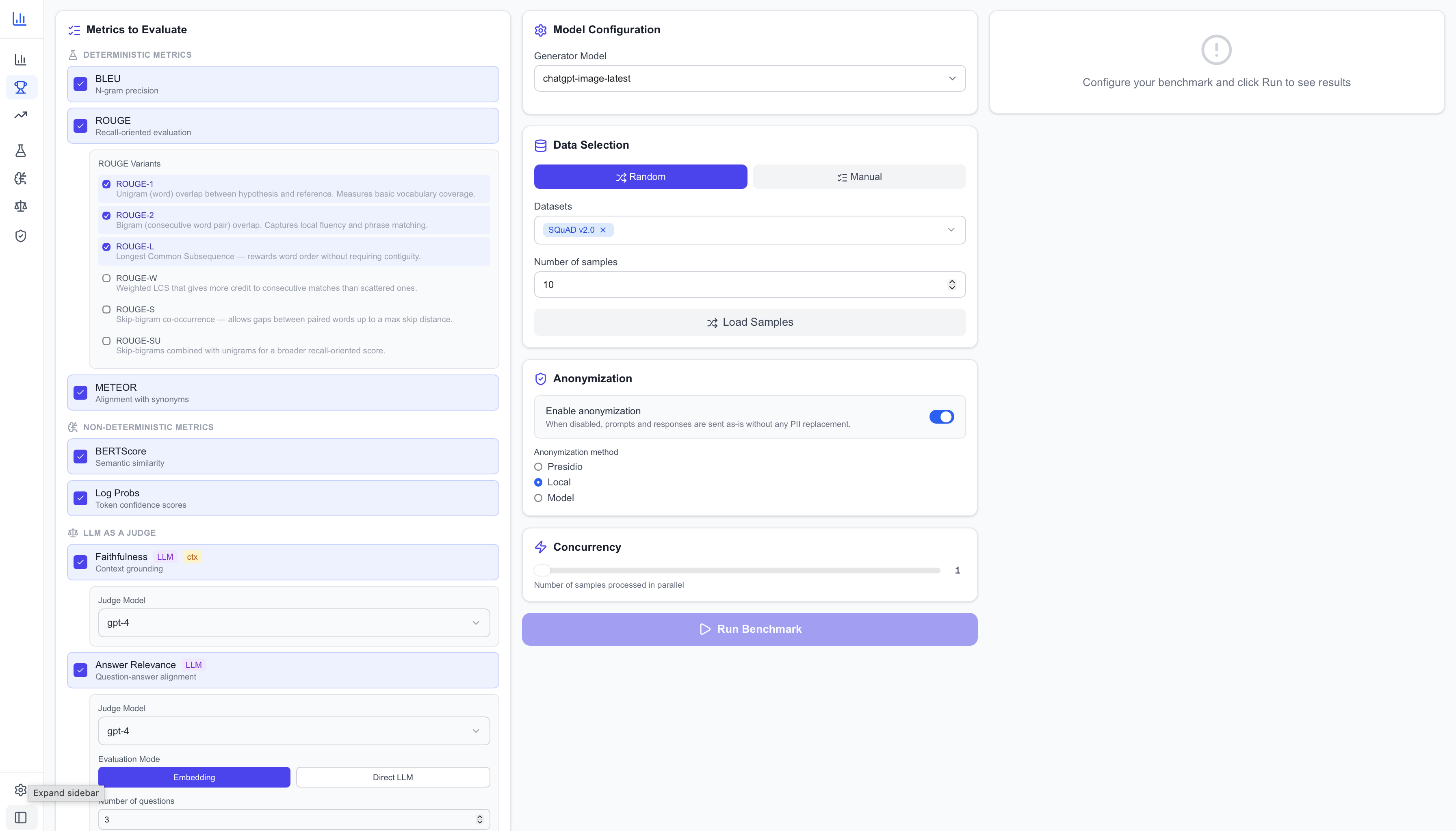}
\caption{Landing page: initializing an LLM evaluation workflow.}
\Description{Screenshot of the LLM-FACETS landing page showing the initialization of an LLM evaluation workflow.}
\label{fig:landing}
\end{figure}
\begin{figure}[h]
\centering
\includegraphics[width=\linewidth]{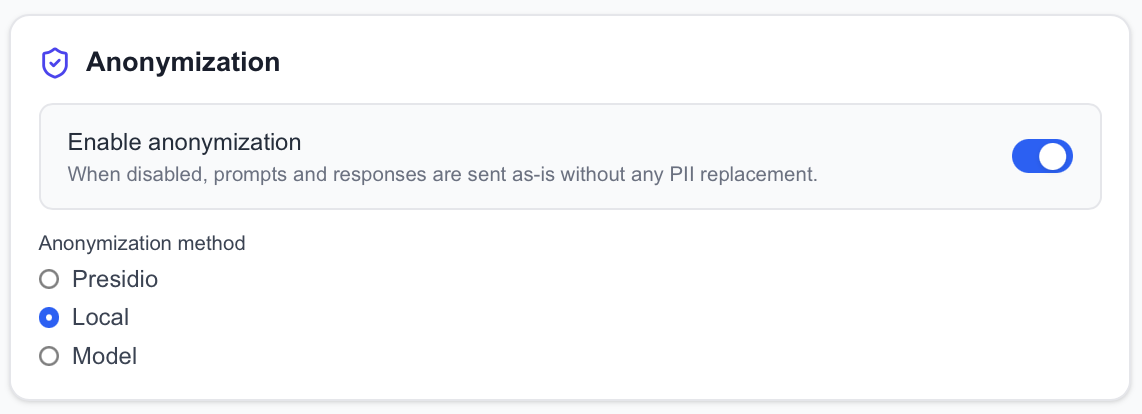}
\caption{Configuration UI for defining practitioner profiles and metric strategies.}
\Description{Screenshot of the LLM-FACETS configuration user interface for defining practitioner profiles and selecting metric strategies.}
\label{fig:strategy}
\end{figure}
\begin{figure*}[t]
  \centering
  \includegraphics[width=\linewidth]{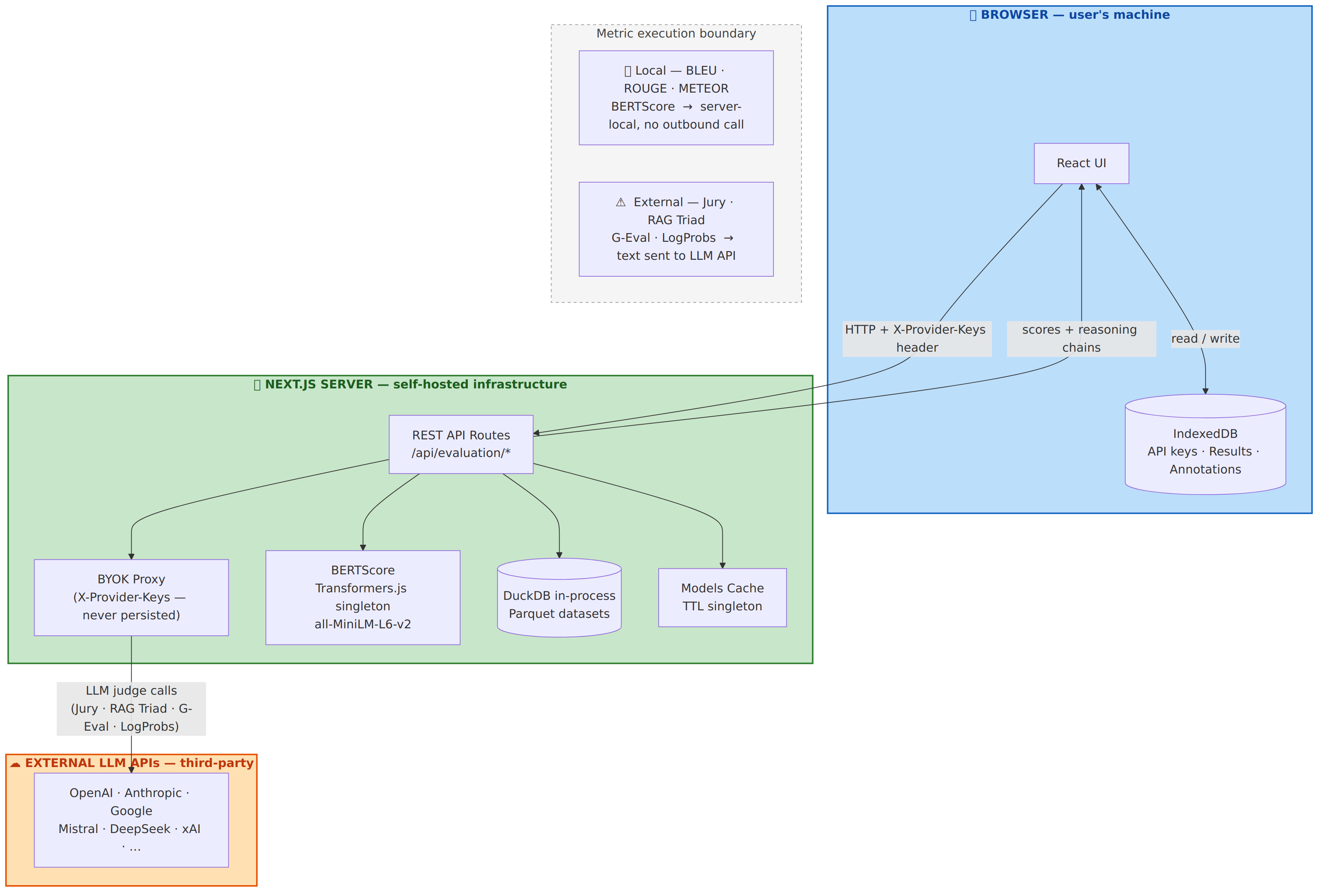}
  \caption{Architecture of LLM-FACETS. \textit{Blue zone}: browser client (IndexedDB stores API keys and results locally). \textit{Green zone}: self-hosted Next.js server process; all computation stays within user infrastructure---BLEU, ROUGE, METEOR, and BERTScore produce no outbound calls. \textit{Orange zone}: external LLM APIs, contacted only for LLM-based metrics (Jury, RAG Triad, G-Eval, LogProbs); data exposure is mitigated by the anonymization pipeline (Section~\ref{sec:architecture}).}
  \Description{Architecture diagram of LLM-FACETS showing three zones: the browser client layer containing the React UI and IndexedDB storage for API keys and results; the self-hosted Next.js server layer containing the REST API routes, BYOK proxy, BERTScore via Transformers.js, DuckDB in-process, and Models Cache; and the external LLM APIs zone contacted only for judge-based metrics.}
  \label{fig:architecture}
\end{figure*}

LLM-FACETS is a Next.js application comprising two layers: a browser client layer (UI, IndexedDB storage, client-side computation) and a self-hosted, stateless server process (API routes, in-process evaluation, BYOK proxy), as illustrated in Figure~\ref{fig:architecture}. The browser client (blue zone) manages the user interface, API key storage via IndexedDB, and evaluation results; the server process (green zone) hosts the REST API routes, the BYOK credential proxy, in-process metric computation (BERTScore via Transformers.js, DuckDB for dataset queries), and the models cache; the external LLM APIs (orange zone) are contacted only when the user explicitly selects LLM-based metrics. The server never persists user data or credentials between requests. Figure~\ref{fig:landing} shows the landing page from which a practitioner initializes an evaluation workflow, and Figure~\ref{fig:strategy} shows the configuration interface for defining practitioner profiles and metric strategies.
The architectural decisions are driven by the challenges established in Section~\ref{sec:challenges}: the interface must be accessible without programming expertise~(C1, C2), data flows must be explicit and controllable~(C3), and metric implementations must be cross-validated and auditable~(C4). This section describes the tool's architecture and the design choices that address these challenges; Section~\ref{sec:metrics} details how the implemented metrics operationalize the transparency dimensions~(C5); Section~\ref{sec:verificationperf} provides the implementation verification and performance evidence.
Two properties are targeted as architectural guarantees of the framework $\mathcal{F}$---intended to hold independently of configuration or deployment context.
\textbf{Data sovereignty.} For deterministic metrics---those whose output is fully determined by the input without stochastic LLM calls---($M_{\mathrm{lex}}$, $M_{\mathrm{neu}}$), all computation executes either client-side via WebAssembly or within the self-hosted Next.js server process; no evaluation data leaves the practitioner's infrastructure. For LLM-judge metrics that require external API calls, the practitioner explicitly supplies credentials via per-request credential injection (the BYOK model detailed in Section~\ref{sec:byok}), and neither credentials nor evaluation data are persisted server-side between requests. Table~\ref{tab:execmodel} in Section~\ref{sec:metrics} documents the execution boundary for each metric.
\textbf{Implementation correctness.} All deterministic metric implementations are cross-validated against canonical reference libraries (Section~\ref{sec:verification}), ensuring numerical reproducibility.
\textbf{Methodological auditability.} LLM-judge metrics expose full reasoning chains in every exported artifact, enabling independent reproduction of any element of $\mathcal{E}$.
\subsection{The BYOK Privacy Model}
\label{sec:byok}
The Bring Your Own Key (BYOK) model is the cornerstone of the framework's privacy guarantees, directly addressing the compliance challenge~(C3). LLM API keys are stored exclusively in the user's browser via IndexedDB---a client-side storage API with origin isolation enforced by the browser security model. Keys are transmitted to LLM-FACETS' server routes using the \texttt{X-Provider-Keys} HTTP header and are used exclusively to instantiate provider objects for the duration of a single request. They are never written to server-side logs, databases, environment files, or any persistent storage. This design means that even a fully compromised server cannot retroactively expose user credentials or reconstruct which keys were used for which evaluations.
For shared-instance deployments (e.g., organizational intranet servers with a single shared API key), the BYOK model falls back to server-side environment variables, maintaining the zero-logging guarantee while enabling centralized key management.
\subsection{In-Process Computation for Data Sovereignty}
Beyond key management, the framework minimizes server-side data retention through two mechanisms, extending the compliance guarantees~(C3) to the computation level:
\textbf{In-process semantic evaluation.} BERTScore computation uses contextual embeddings (dense vector representations capturing semantic meaning) generated via Transformers.js~\cite{transformersjs} within the self-hosted Next.js server process. The embedding model (\texttt{all-MiniLM-L6-v2}) is loaded once as a server-side singleton and processes all inputs locally, without transmitting text or embeddings to any external service. This enables organizations to perform neural similarity evaluation~\cite{zhang2019bertscore} on sensitive documents (e.g., clinical notes, legal briefs) while keeping all data within their own infrastructure.
\textbf{In-process dataset analysis.} Dataset management uses DuckDB~\cite{raasveldt2019duckdb}, an in-process analytical database engine that reads Apache Parquet files~\cite{parquet2013} directly without requiring a server process. Datasets are stored locally under a configurable path. All queries (schema introspection, random sampling, SQL filtering) execute within the Next.js server process and are never transmitted to external services. This architecture enables SQL-level analysis of large evaluation corpora while preserving full data sovereignty~(C3) and enabling reproducible dataset exploration without external tooling~(C1).
\subsection{Privacy-Preserving Input Anonymization}
To evaluate LLM outputs on corpora containing personally identifiable information (PII)---a prerequisite for auditing in healthcare, legal, and financial domains---LLM-FACETS provides a configurable anonymization pipeline that redacts sensitive entities before any text is transmitted to an external LLM judge API. This extends the data sovereignty guarantee of the BYOK architecture to the content level: identifying information is eliminated from the evaluated text prior to any outbound network call.
The pipeline operates in two sequential stages. A first stage deterministically applies regular expressions to identify four high-precision structural entity types: email addresses, telephone numbers (using standardized regex patterns), IPv4 addresses, and credit card numbers. A second, pluggable stage handles linguistic entities through Named Entity Recognition (NER), with three configurable strategies:
\textbf{Local NER (compromise.js~\cite{compromisejs}).} A server-side JavaScript NLP library, integrated as a dependency within the Next.js server process (green zone in Figure~\ref{fig:architecture}), that detects PERSON, LOCATION, and ORGANIZATION entities using grammatical rules, with no machine learning model and no network call. This strategy operates fully offline with negligible latency ($< 10$~ms per 1K words) and is natively available within the Node.js environment without external service dependencies.
\textbf{Presidio~\cite{presidio2018}.} A standalone self-hosted NER service supporting over seven entity types across more than 20 languages with configurable confidence thresholds. When deployed as an adjacent microservice on an organizational intranet, it provides high-precision multilingual detection without transmitting data to any third party, satisfying the same GDPR Article~28 guarantees as the BYOK model.
\textbf{Model-based NER~\cite{gemini2024}.} A Gemini LLM configured at temperature $= 0$ (a parameter controlling output randomness; setting it to zero produces the most deterministic, reproducible responses) returns a structured JSON array of detected entities, providing context-aware detection of complex co-references and domain-specific identifiers that rule-based approaches may miss.
Across all strategies, the internal \texttt{EntityTracker} component automatically applies to enforce \textit{co-reference consistency}: each unique entity is assigned a typed, numbered placeholder upon first occurrence (e.g., \texttt{[PERSON\_1]}, \texttt{[LOCATION\_2]}), and the same placeholder is reused for all subsequent occurrences within the document. This preserves the referential structure of the text---semantically meaningful for evaluating factual coherence---while eliminating identifying information. The active strategy is selectable per-session through the \toolname{} interface; administrators may enforce a fixed strategy at the server level via environment variable, enabling policy-compliant deployment in regulated environments.
It is important to note that metrics are computed over the anonymized text, not the original. Lexical overlap metrics (BLEU, ROUGE, METEOR) will show reduced scores when evaluated against non-anonymized reference texts, since placeholders replace named entities. This trade-off is expected: the anonymization pipeline prioritizes data protection at the cost of absolute metric fidelity. Practitioners should ensure reference texts are anonymized consistently when using the anonymization pipeline for metric-grounded evaluation.
\subsection{Plugin Architecture: Adding Metrics, Providers, and Datasets}
\begin{figure}[t]
  \centering
  \includegraphics[width=\linewidth]{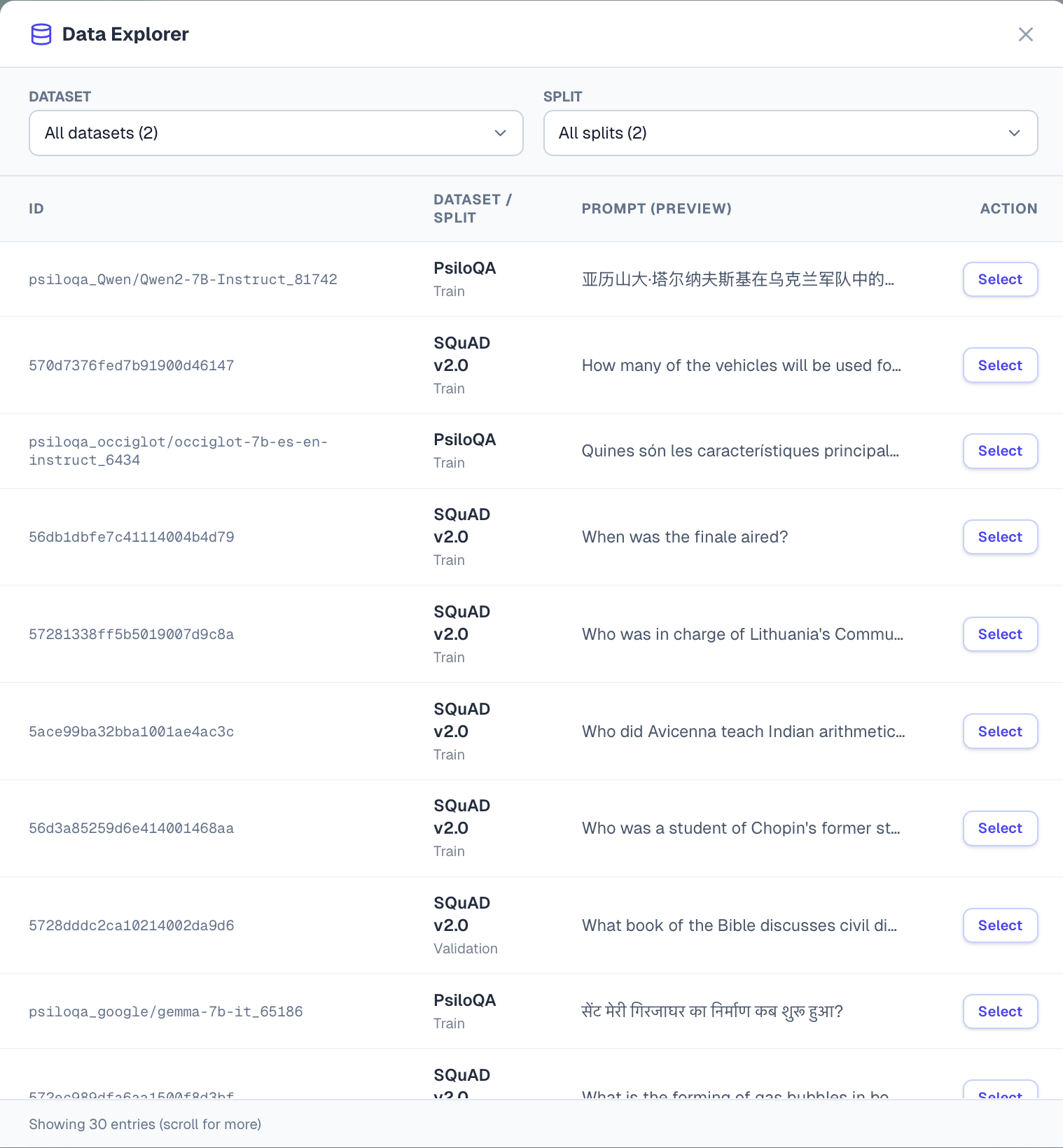}
  \caption{Dataset explorer interface showing available datasets (SQuAD~v2, PsiloQA), download status, split selection, and live row sampling with DuckDB results.}
  \Description{Screenshot of the dataset selector modal in LLM-FACETS, showing two dataset cards for SQuAD v2 and PsiloQA with download progress indicators, split selectors, and a table of sampled rows.}
  \label{fig:dataset}
\end{figure}

The framework is built on Next.js~16 (App Router), which provides server-side rendering and a RESTful API layer. Four design patterns underpin its extensibility. They map directly to the Next.js server components shown in Figure~\ref{fig:architecture}. Together they ensure that adding a new capability to LLM-FACETS never requires modifying the core evaluation pipeline.
\textbf{Metric Registry.} Located in the API Routes layer of the server process (green zone in Figure~\ref{fig:architecture}), all metrics implement a common \texttt{Metric} interface exposing a \texttt{compute(sample)} function that returns a uniformly normalized score in $[0, 1]$ (where 1 consistently indicates the optimal or most-reliable result regardless of the underlying metric semantics, meaning a low score always signifies poor quality or high hallucination risk), together with optional reasoning chains and intermediate outputs. Adding a new metric requires three steps: (1)~implement the \texttt{Metric} interface; (2)~register it with metadata (name, description, category, transparency dimension) in the central registry; (3)~no further changes are needed---the metric is automatically available in the \emph{Navigation} and \emph{React UI Components} (Figure~\ref{fig:architecture}), benchmarking dashboard, REST API, and export artifacts. This means any metric proposed in future research can be integrated without modifying the evaluation pipeline.
\textbf{Dataset Registry.} Implemented within the Dataset Service (Figure~\ref{fig:architecture}), datasets extend an abstract \texttt{DatasetBase} class that exposes a single \texttt{normalize(row)} method mapping dataset-specific field names to the uniform \texttt{DatasetSample} interface. Adding a new dataset requires implementing \texttt{normalize(row)} and registering the dataset metadata; it then becomes available in the dataset explorer (Figure~\ref{fig:dataset}), benchmarking pipeline, and sample viewer. Four datasets are currently implemented: SQuAD~v2~\cite{rajpurkar2018squad} and PsiloQA~\cite{psiloqa2024} for reading-comprehension Question Answering (QA), SelfAware~\cite{yin2023selfaware} for knowledge-boundary and refusal evaluation, and HaluEval~\cite{li2023halueval} for faithfulness and correctness evaluation with explicit hallucination labels.
\textbf{Provider Factory Pattern.} Located in the Provider Factory component (Figure~\ref{fig:architecture}), all LLM interactions are mediated through an abstract \texttt{LLMProvider} base class. A factory map associates provider names (OpenAI, Anthropic, Google Gemini, DeepSeek, Groq, Mistral, xAI, MiniMax, Devana, Z.ai) with constructor functions. Each provider implements \texttt{call()} for text generation and \texttt{getAvailableModels()} for dynamic model listing. These providers are only contacted when the user explicitly selects an LLM-judge metric and supplies their own API key; deterministic metrics require no provider configuration.
\textbf{Server-side caching.} A \texttt{ModelsCache} singleton (Figure~\ref{fig:architecture}) with a configurable TTL (default: 1~hour) caches available model lists per provider, reducing redundant API calls and improving responsiveness.
\subsection{Statistical Benchmark Analysis Dashboard \textnormal{\normalfont[Process]}}
\begin{figure}[t]
  \centering
  \includegraphics[width=\linewidth]{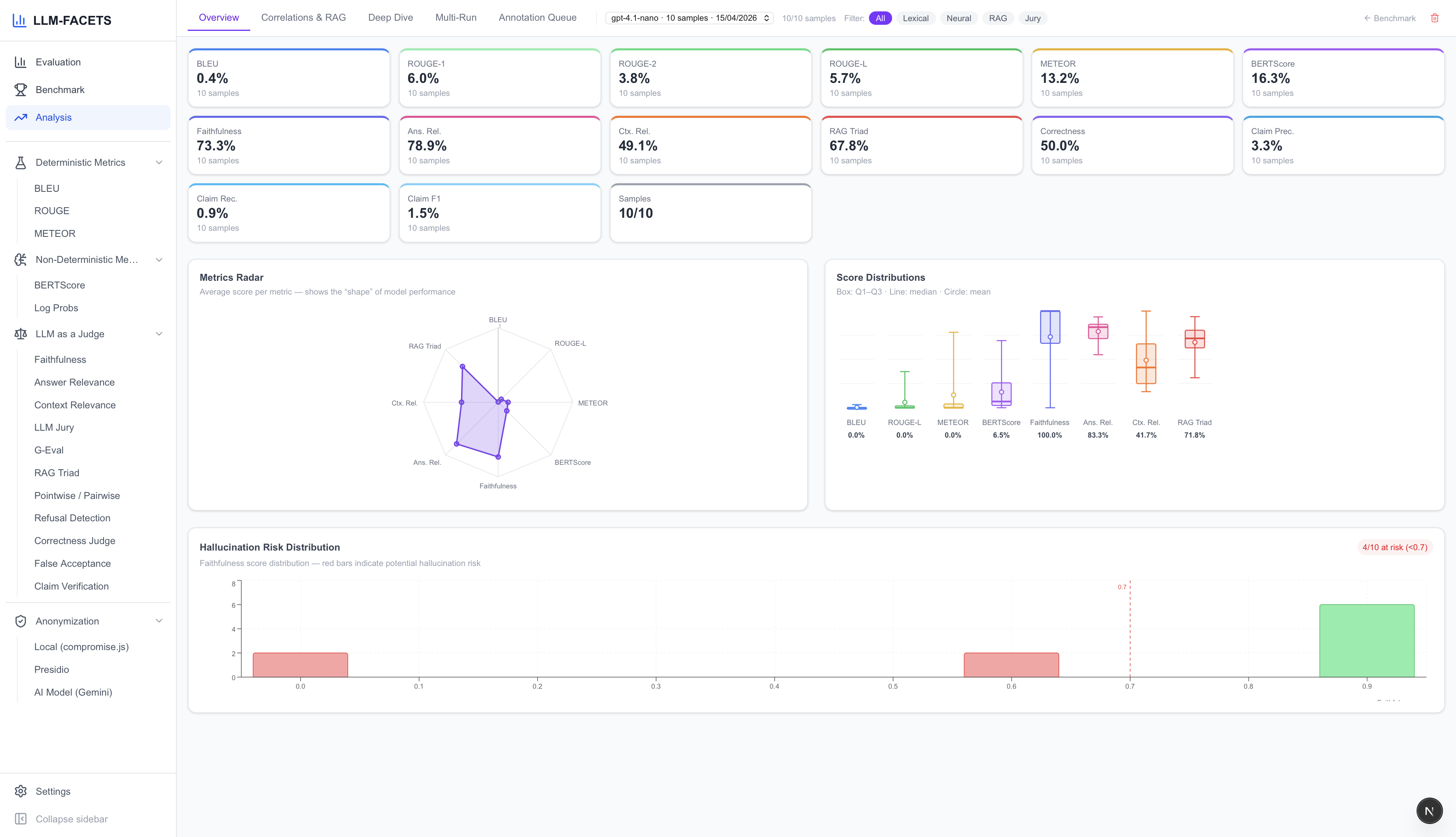}
  \caption{Benchmark analysis dashboard (Overview tab): radar chart aggregating nine primary metrics, box plots showing per-metric score distributions, and hallucination risk histogram. All visualizations are generated client-side from IndexDB-stored results without any server-side data transmission.}
  \Description{Screenshot of the benchmark analysis overview tab in LLM-FACETS. A radar chart in the top-left displays nine metric axes. Box plots on the right show score distributions for each metric with quartile markers. A hallucination histogram at the bottom bins faithfulness scores and highlights bins below 0.7 in red.}
  \label{fig:analysis}
\end{figure}

Beyond per-sample metric evaluation, LLM-FACETS includes a multi-tab analysis dashboard that derives statistical insights from complete benchmark runs (Figure~\ref{fig:analysis}). Results are persisted in the browser's IndexedDB store and loaded directly into the analysis view without any server-side storage, preserving the zero-transmission guarantee. Four analytical views are provided, addressing the accessibility~(C2) and reproducibility~(C4) challenges, with metrics filterable by category (Lexical, Neural, RAG, Jury introduced in Section~\ref{sec:metrics}):
A radar chart aggregates median values across nine primary metrics (BLEU, ROUGE-L, METEOR, BERTScore, Faithfulness, Answer Relevance, Context Relevance, RAG Triad composite, and Jury Score), providing a compact multi-dimensional quality signature for a benchmark run. Box plots display the full quartile distribution ($Q_1$, median, $Q_3$, whiskers at $\mathrm{min}/\mathrm{max}$, and a mean marker) for each metric, exposing score spread and outliers that aggregate statistics conceal. A dedicated hallucination histogram bins faithfulness scores into ten equal intervals and flags bins below 0.7 as elevated-risk distributions, providing at-a-glance triage of potential hallucination prevalence.
A Pearson correlation heatmap (visible in the Correlation tab of the dashboard, not shown in Figure~\ref{fig:analysis}) displays the $N \times N$ inter-metric correlation matrix, computed as:
\[
  r_{XY} = \frac{\displaystyle\sum_{i} (x_i - \bar{x})(y_i - \bar{y})}{\sqrt{\displaystyle\sum_{i}(x_i - \bar{x})^2 \cdot \displaystyle\sum_{i}(y_i - \bar{y})^2}}
\]
where $r_{XY}$ is the correlation coefficient between two metrics $X$ and $Y$, $x_i$ and $y_i$ are the individual scores for the $i$-th sample, and $\bar{x}, \bar{y}$ are the respective mean scores. Cell color intensity encodes correlation strength (blue for positive, red for negative). Three pairwise RAG scatter plots (Faithfulness vs.\ Answer Relevance, Faithfulness vs.\ Context Relevance, and Answer Relevance vs.\ Context Relevance) expose structural dependencies between the RAG Triad components. A jury agreement histogram shows the distribution of inter-judge consensus across samples, providing an aggregate view of evaluator stability.
\textbf{Deep Dive.} A paginated per-sample heatmap renders a color-coded grid of all metric values across samples (30 per page), enabling rapid visual identification of underperforming individual outputs. A BERTScore precision-versus-recall scatter plot, with points sized by F1 score, diagnoses whether semantic similarity shortfalls originate from precision deficits (hallucinated content) or recall deficits (missing information). An execution timeline tracks up to five metric trajectories across the sample execution order, exposing performance drift over the course of a benchmark run.
\textbf{Multi-Run Comparison.} A comparative radar chart and a grouped bar chart overlay up to five benchmark runs simultaneously on the same axes, enabling direct comparison of model versions, prompt strategies, or provider configurations evaluated on identical corpora.
This dashboard transforms raw benchmark data into auditable statistical evidence legible to non-technical practitioners without recourse to scripting or external analytical tooling, operationalizing the process transparency requirement articulated in Section~\ref{sec:transparencydimensions}.
\begin{figure}[t]
  \centering
  \includegraphics[width=\linewidth]{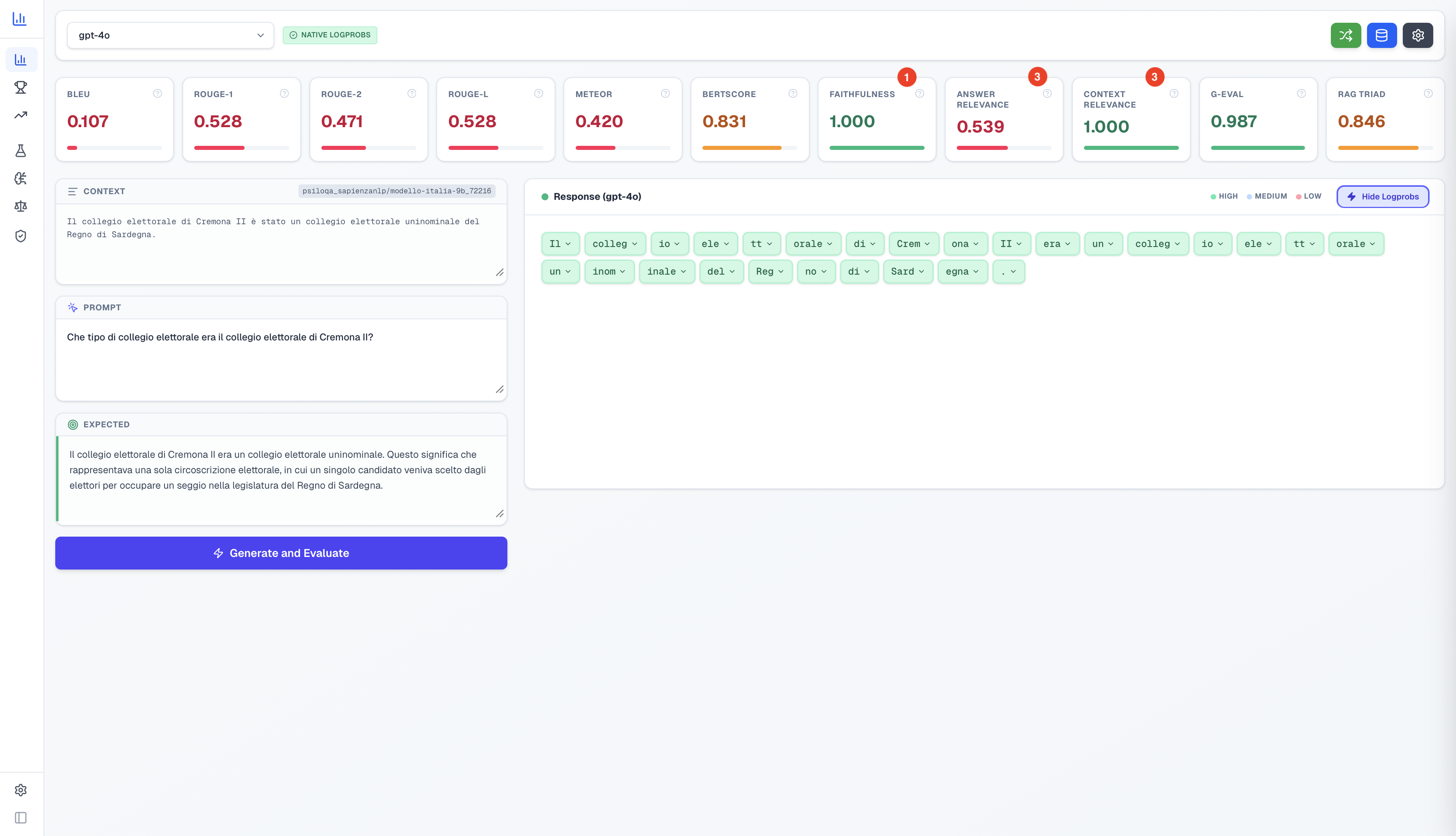}
  \caption{Main dashboard showing the metrics selection grid. Users can navigate between categories (Traditional, Neural, LLM-as-a-Judge, RAG) and launch evaluations directly.}
  \Description{Screenshot of the \toolname{} dashboard showing a grid of metric cards organized in four categories: Traditional (BLEU, ROUGE, METEOR), Neural (BERTScore, LogProbs), LLM-as-a-Judge (G-Eval, Jury, Evaluation Topologies), and RAG (Faithfulness, Answer Relevance, Context Relevance).}
  \label{fig:dashboard}
\end{figure}
\section{Transparency Evaluation Metrics}
\label{sec:metrics}
The tool architecture described in Section~\ref{sec:architecture} provides the infrastructure for privacy-preserving, extensible evaluation. This section details how the implemented metrics operationalize the four transparency dimensions identified in Section~\ref{sec:transparencydimensions}---baseline grounding, epistemic, factual, and process---each addressing a distinct practitioner question and directly responding to the transparency challenge~(C5). Table~\ref{tab:metrics} provides a complete inventory of all implemented metrics with their categories, execution types, and transparency dimensions. Table~\ref{tab:execmodel} clarifies where each metric executes, making explicit the data sovereignty boundary~(C3) for organizations operating under GDPR, HIPAA, or similar data protection regimes: \textit{Next.js server-local} means the computation runs within the self-hosted server process; no text or embeddings are transmitted to any external service.
\begin{table}[h]
  \caption{Execution location for each metric category. ``Next.js server-local'' means the computation is confined to the self-hosted server process. Only Jury and RAG Triad metrics require outbound calls to external LLM APIs.}
  \label{tab:execmodel}
  \small
  \begin{tabular}{ll}
    \toprule
    \textbf{Metric} & \textbf{Execution location} \\
    \midrule
    BLEU, ROUGE, METEOR     & Next.js server-local (pure computation) \\
    BERTScore               & Next.js server-local (Transformers.js singleton) \\
    LogProbs                & External LLM API (provider-native) \\
    G-Eval, Eval Topologies & External LLM API \\
    Jury                    & External LLM APIs (one call per judge) \\
    Faithfulness, AR, CR    & External LLM API (RAG chain) \\
    \bottomrule
  \end{tabular}
\end{table}
\subsection{Auditing Hallucinations: The RAG Triad \textnormal{\normalfont[Factual]}}
The RAG Triad (Faithfulness, Answer Relevance, Context Relevance, detailed in Table~\ref{tab:ragtriad}) implements factual transparency checks for RAG pipelines, operationalizing the three questions introduced in Section~\ref{sec:transparencydimensions}: \textit{Is the answer grounded in the retrieved context? Does the answer address the query? Is the retrieved context relevant to the query?}
\begin{table}[h]
  \caption{RAG Triad Components}
  \label{tab:ragtriad}
  \small
  \begin{tabular}{llll}
    \toprule
    \textbf{Component} & \textbf{Method} & \textbf{Output} & \textbf{Score Range} \\
    \midrule
    Faithfulness & CoT claim decomposition & Supported claims ratio & $[0, 1]$ + specific unsupported claims \\
    Answer Relevance & Reverse question generation & Cosine similarity & $[0, 1]$ \\
    Context Relevance & Sentence classification & Signal/noise ratio & $[0, 1]$ \\
    \bottomrule
  \end{tabular}
\end{table}
By presenting all three scores simultaneously with per-claim breakdowns (Figure~\ref{fig:ragtriad}), the interface allows a domain expert without programming skills to trace a hallucination back to its source in the retrieval pipeline, directly addressing the onboarding barrier~(C1).
\subsection{Visualizing Model Confidence: Token-Level Log-Probabilities \textnormal{\normalfont[Epistemic]}}
\label{sec:logprobs}
LogProb visualization surfaces epistemic transparency. For providers that return per-token log-probabilities (OpenAI, DeepSeek, xAI), LLM-FACETS renders the generated text with each token color-coded across five confidence tiers (described in Table~\ref{tab:confidence}), translating the mathematical concept of output probability $p(t_i \mid t_{<i})$ into an interface accessible to non-technical auditors:
\begin{table}[h]
  \caption{Confidence tiers and corresponding log-probability thresholds}
  \label{tab:confidence}
  \small
  \begin{tabular}{llc}
    \toprule
    \textbf{Confidence Tier} & \textbf{Color Code} & \textbf{Probability Threshold ($p(t_i \mid t_{<i})$)} \\
    \midrule
    Very high & Dark green & $p \geq 0.8$ \\
    High & Light green & $0.6 \leq p < 0.8$ \\
    Medium & Yellow & $0.4 \leq p < 0.6$ \\
    Low & Orange & $0.2 \leq p < 0.4$ \\
    Very low & Red & $p < 0.2$ \\
    \bottomrule
  \end{tabular}
\end{table}
Hovering over any token reveals its exact log-probability and the top-$k$ alternative tokens the model considered.
This visualization serves as an accessibility layer for uncertainty quantification: auditors who are not familiar with log-probability mathematics can immediately identify when a model is ``guessing'' versus when it expresses strong confidence. In combination with the RAG Triad, this allows correlation between low-confidence tokens and unfaithful claims---a powerful diagnostic for systematic hallucination analysis.
\subsection{Mitigating Judge Bias: Multi-Judge Consensus \textnormal{\normalfont[Process]}}
\begin{figure}[t]
  \centering
  \includegraphics[width=\linewidth]{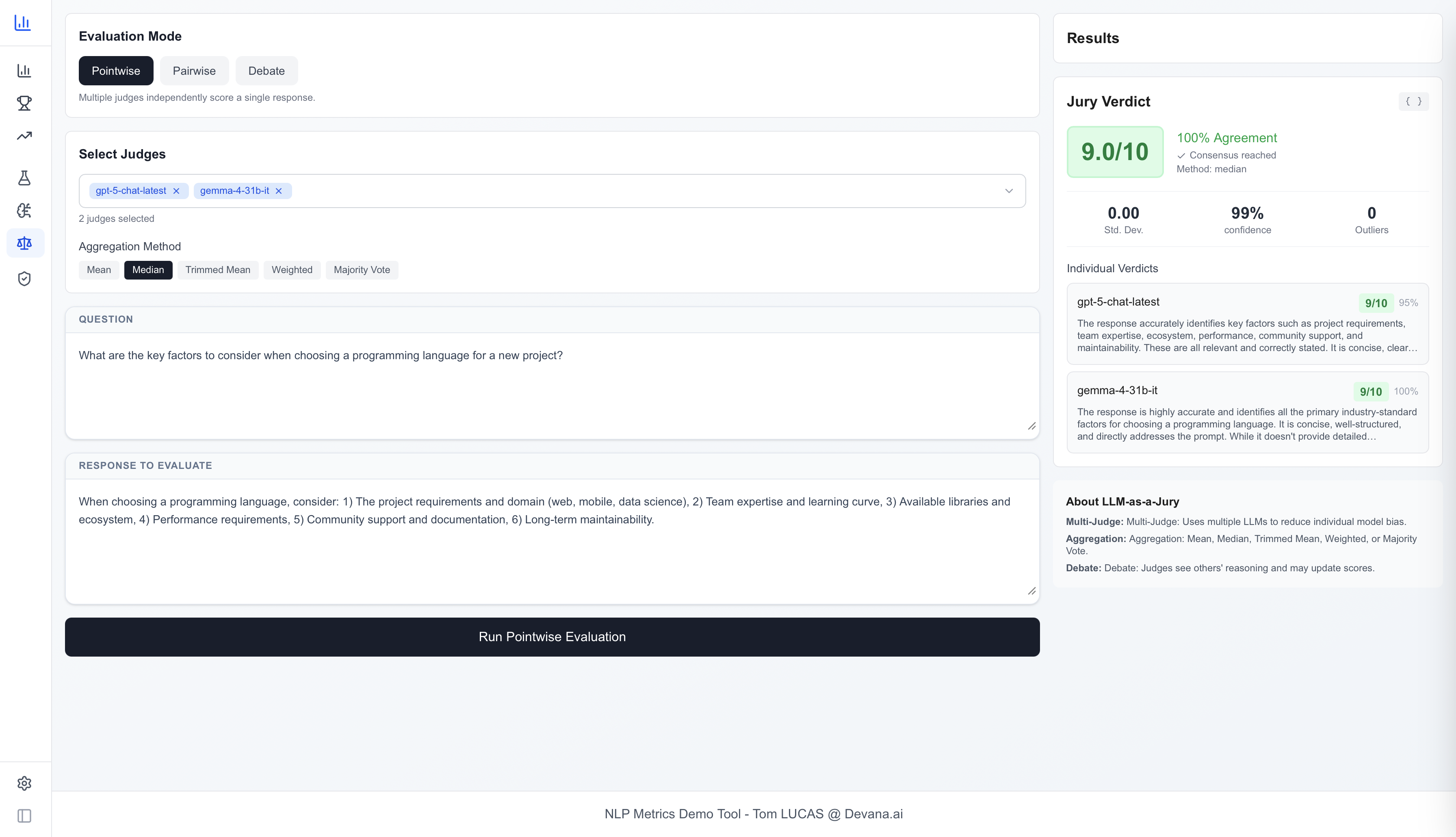}
  \caption{Jury module: three judges from different provider families (OpenAI, Google, Alibaba) independently score the same input using pointwise evaluation with median aggregation. Individual scores, reasoning chains, inter-judge agreement ($\sigma$), and outlier flags are reported simultaneously.}
  \Description{Screenshot of the Jury evaluation interface in LLM-FACETS. Three judge panels show individual scores and reasoning from gpt-4o-mini, gemini-flash, and qwen3-32b. A summary panel displays the median consensus score, standard deviation, and agreement index.}
  \label{fig:jury}
\end{figure}

The Jury module implements process transparency by exposing the unreliability of single-judge LLM evaluation (Figure~\ref{fig:jury}). A configurable set of $n$ judges (potentially from different model providers and families) independently score the same input. The tool reports:
\begin{itemize}
  \item Individual judge scores and their reasoning chains.
  \item Consensus score (mean or median, configurable).
  \item Inter-judge agreement ($\sigma$ of scores); high variance signals contested assessments.
  \item Statistical outlier detection to identify rogue judges.
\end{itemize}
The Evaluation Topologies module extends this with pairwise comparative evaluation ($O(N^2)$ or tournament-style) where position-bias mitigation is achieved by running each comparison in both orderings (A vs.\ B and B vs.\ A) and averaging. This directly addresses the single largest known bias in LLM-as-a-Judge evaluation~\cite{zheng2023judging}.
This multi-judge design is empirically grounded in recent work. The Panel of LLM Evaluators (PoLL)~\cite{poll2024} demonstrates that a diverse jury of smaller, heterogeneous models outperforms a single frontier judge across multiple evaluation datasets while reducing cost by up to $7\times$---a finding that validates the Jury module's default configuration using models from different provider families. At the aggregation level, CARE~\cite{care2026} models inter-judge correlations to isolate shared confounders---response length, stylistic similarity---from true quality signals, providing a statistically principled estimator. LLM-FACETS' trimmed-mean and weighted aggregation strategies address the same problem at lower computational cost, prioritizing deployment accessibility over statistical optimality while exposing inter-judge variance as a first-class reported statistic.
\subsection{Baseline Grounding: Traditional and Neural Metrics \textnormal{\normalfont[Baseline Grounding]}}
Traditional n-gram metrics (BLEU~\cite{papineni2002bleu}, ROUGE~\cite{lin2004rouge}, METEOR~\cite{banerjee2005meteor}) and neural similarity (BERTScore~\cite{zhang2019bertscore}) provide deterministic, reproducible baselines that anchor LLM-FACETS in well-understood, externally validated methods.
These baselines serve a specific role in baseline grounding transparency (Section~\ref{sec:transparencydimensions}): they provide deterministic, cross-validated reference measurements that allow auditors to detect systematic divergences between traditional metric scores and LLM judge scores, which may indicate prompt engineering artifacts, model self-bias, or genuine quality differentials that word-overlap metrics cannot capture.
Table~\ref{tab:metrics} summarizes all implemented metrics across the four categories (Traditional, Neural, LLM-as-a-Judge, RAG). Note that the 18~metric \textit{variants} reported in Section~\ref{sec:conclusion} arise from multi-variant metrics: ROUGE alone yields six variants (ROUGE-1, ROUGE-2, ROUGE-L, ROUGE-W, ROUGE-S, ROUGE-SU), and BERTScore yields three (Precision, Recall, F1).
\begin{table*}
  \caption{Summary of implemented evaluation metrics in LLM-FACETS.}
  \label{tab:metrics}
  \begin{tabular}{llll}
    \toprule
    \textbf{Metric} & \textbf{Category} & \textbf{Type} & \textbf{Transparency Dimension} \\
    \midrule
    BLEU~\cite{papineni2002bleu} & Traditional & Deterministic & Baseline grounding \\
    ROUGE~\cite{lin2004rouge} & Traditional & Deterministic & Baseline grounding \\
    METEOR~\cite{banerjee2005meteor} & Traditional & Deterministic & Baseline grounding \\
    BERTScore~\cite{zhang2019bertscore} & Neural & Deterministic (server-local) & Baseline grounding \\
    LogProbs & Neural & Provider-native & Epistemic transparency \\
    G-Eval~\cite{liu2023geval} & LLM-as-a-Judge & LLM-based & Process transparency \\
    Evaluation Topologies & LLM-as-a-Judge & LLM-based & Process transparency \\
    Jury & LLM-as-a-Judge & LLM-based (multi-judge) & Process transparency \\
    Faithfulness~\cite{ragas2024} & RAG & LLM-based & Factual transparency \\
    Answer Relevance~\cite{ragas2024} & RAG & LLM-based & Factual transparency \\
    Context Relevance~\cite{ragas2024} & RAG & LLM-based & Factual transparency \\
    \bottomrule
  \end{tabular}
\end{table*}
\section{Implementation Verification and Performance}
\label{sec:verificationperf}
Credible evaluation requires verified implementations. The reproducibility challenge~(C4) identified in Section~\ref{sec:challenges} demands that metric implementations be cross-validated against established reference libraries and that performance characteristics be documented. This section provides that evidence.
\subsection{Cross-Validation Against Reference Implementations}
\label{sec:verification}
To ensure the reproducibility of our implementations, all metric implementations in LLM-FACETS undergo rigorous cross-validation against established Python reference libraries (e.g., NLTK, Hugging Face \texttt{evaluate}). We executed a comprehensive test suite comprising 267 validation cases. Deterministic metrics---including BLEU, ROUGE variants, METEOR, and BERTScore---passed all tests, matching their Python reference implementations with an absolute error below $10^{-5}$. Table~\ref{tab:bleu_validation} provides evidence of a sample validation, and Table~\ref{tab:verification} reports maximum absolute errors measured over 100~randomly sampled test cases from the SQuAD~v2 validation set.
For non-deterministic LLM-based metrics (RAG Triad, G-Eval, Jury), we evaluate score consistency by computing the standard deviation ($\sigma$) across repeated runs ($n=5$). In our validation suite using the \texttt{gpt-4o-mini} model, all LLM-based metrics maintained $\sigma < 0.05$ across evaluation runs, confirming sufficient score stability for production use.
\begin{table}[h]
  \caption{Validation of TypeScript BLEU implementation against Python \texttt{nltk.translate.bleu\_score} reference on a standardized Wikipedia excerpt.}
  \label{tab:bleu_validation}
  \small
  \begin{tabular}{llrr}
    \toprule
    \textbf{Variant} & \textbf{Direction} & \textbf{NLTK Score} & \textbf{TS Score} \\
    \midrule
    BLEU-1 & Hyp=Context, Ref=Response & $3.52490\%$ & $3.52490\%$ \\
    BLEU-2 & Hyp=Response, Ref=Context & $0.03366\%$ & $0.03366\%$ \\
    \bottomrule
  \end{tabular}
\end{table}
\begin{table}[h]
  \caption{Cross-validation of deterministic metric implementations against Python reference libraries. Maximum absolute error over 100 test cases.}
  \label{tab:verification}
  \small
  \begin{tabular}{llll}
    \toprule
    \textbf{Metric} & \textbf{Reference library} & \textbf{Max $|$error$|$} \\
    \midrule
    BLEU            & NLTK \texttt{bleu\_score}   & $< 10^{-5}$ \\
    ROUGE (all variants) & \texttt{rouge-score}    & $< 10^{-5}$ \\
    METEOR          & NLTK \texttt{meteor\_score} & $< 10^{-5}$ \\
    BERTScore P/R/F & HuggingFace \texttt{evaluate} & $< 10^{-5}$ \\
    \bottomrule
  \end{tabular}
\end{table}
\subsection{Performance Analysis}
\label{sec:performance}
A critical challenge in LLM evaluation is the computational overhead of metric calculation, particularly over large datasets. To address transparency and streamline the process, the tool provides a comprehensive dashboard grid (illustrated in Figure~\ref{fig:dashboard}) from which evaluations are launched. To evaluate the computational efficiency of the LLM-FACETS tool, we conducted a benchmark over the SelfAware~\cite{yin2023selfaware} (1,047 samples) and HaluEval~\cite{li2023halueval} (12,322 samples) datasets on a high-performance server (Intel Xeon Platinum 8480+ with 28~vCPUs, 251GB~RAM, and a single NVIDIA H100 80GB GPU, running Ubuntu 22.04 LTS).
Processing a batch of 13,369 samples from these datasets to compute generations and traditional metrics (BLEU, ROUGE, METEOR, BERTScore) took a total of 24.7 seconds. Out of this total, text generation using the Mistral-Small-3.2-24B-Instruct model (via vLLM) accounted for 17.9 seconds (1.3 ms/sample), while the actual computation of the traditional metrics executed in merely 6.7 seconds (0.5 ms/sample/metric). This provides evidence of the efficiency of our in-process metric computation engine.
\section{Running Examples}
\label{sec:usecases}
\subsection{Example 1: The Developer -- RAG Pipeline Debugging}
\begin{figure}[t]
  \centering
  \includegraphics[width=\linewidth]{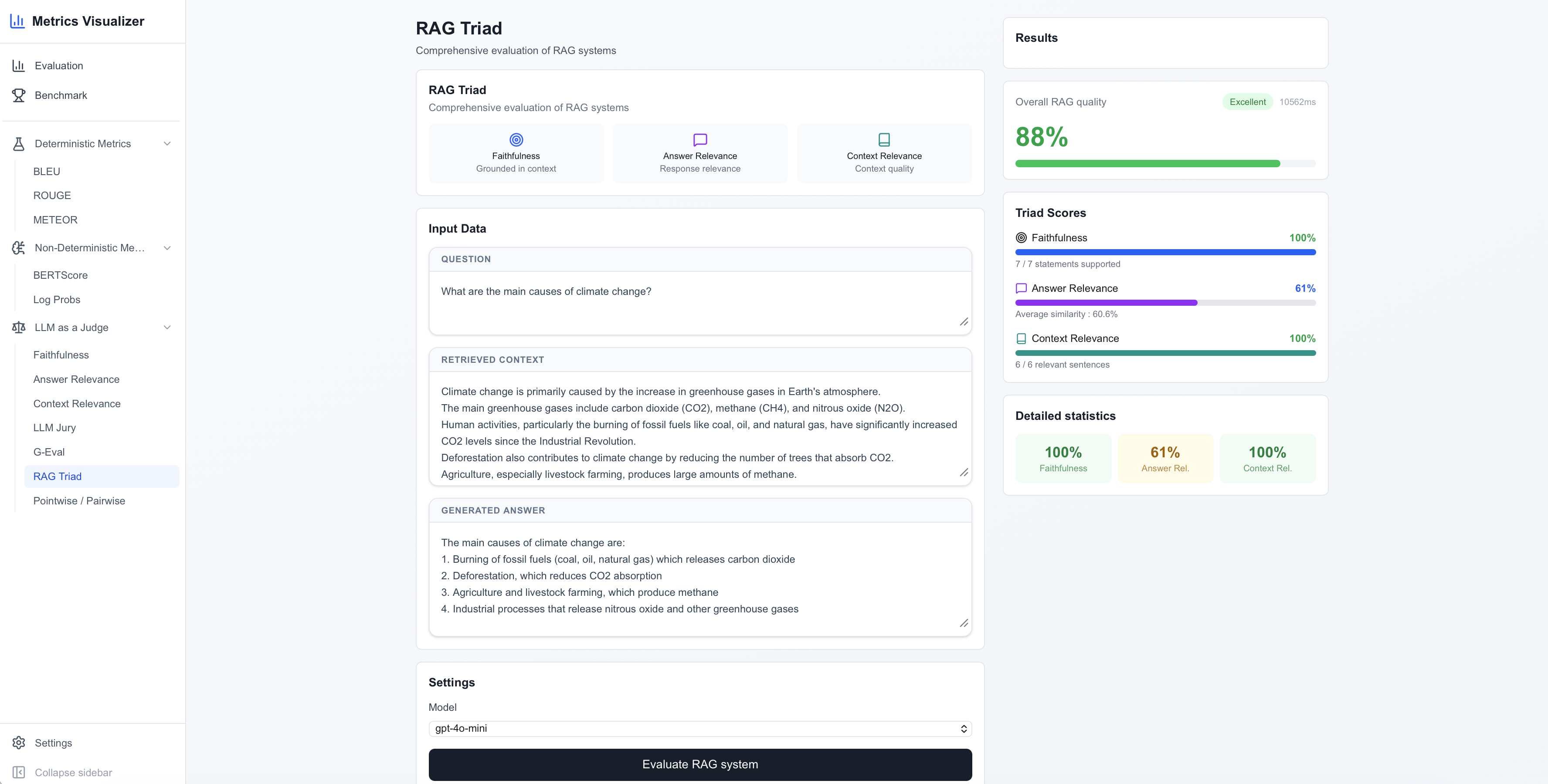}
  \caption{RAG Triad evaluation interface showing a complete example: the user-provided question, retrieved context, and generated answer, alongside the three computed scores.}
  \Description{Screenshot of the RAG Triad interface in LLM-FACETS. The left panel shows the input fields for question, context, and generated answer. The right panel displays three circular score gauges for Faithfulness (0.6), Answer Relevance (0.85), and Context Relevance (0.4), with a per-claim breakdown below.}
  \label{fig:ragtriad}
\end{figure}

\begin{figure}[t]
  \centering
  \includegraphics[width=\linewidth]{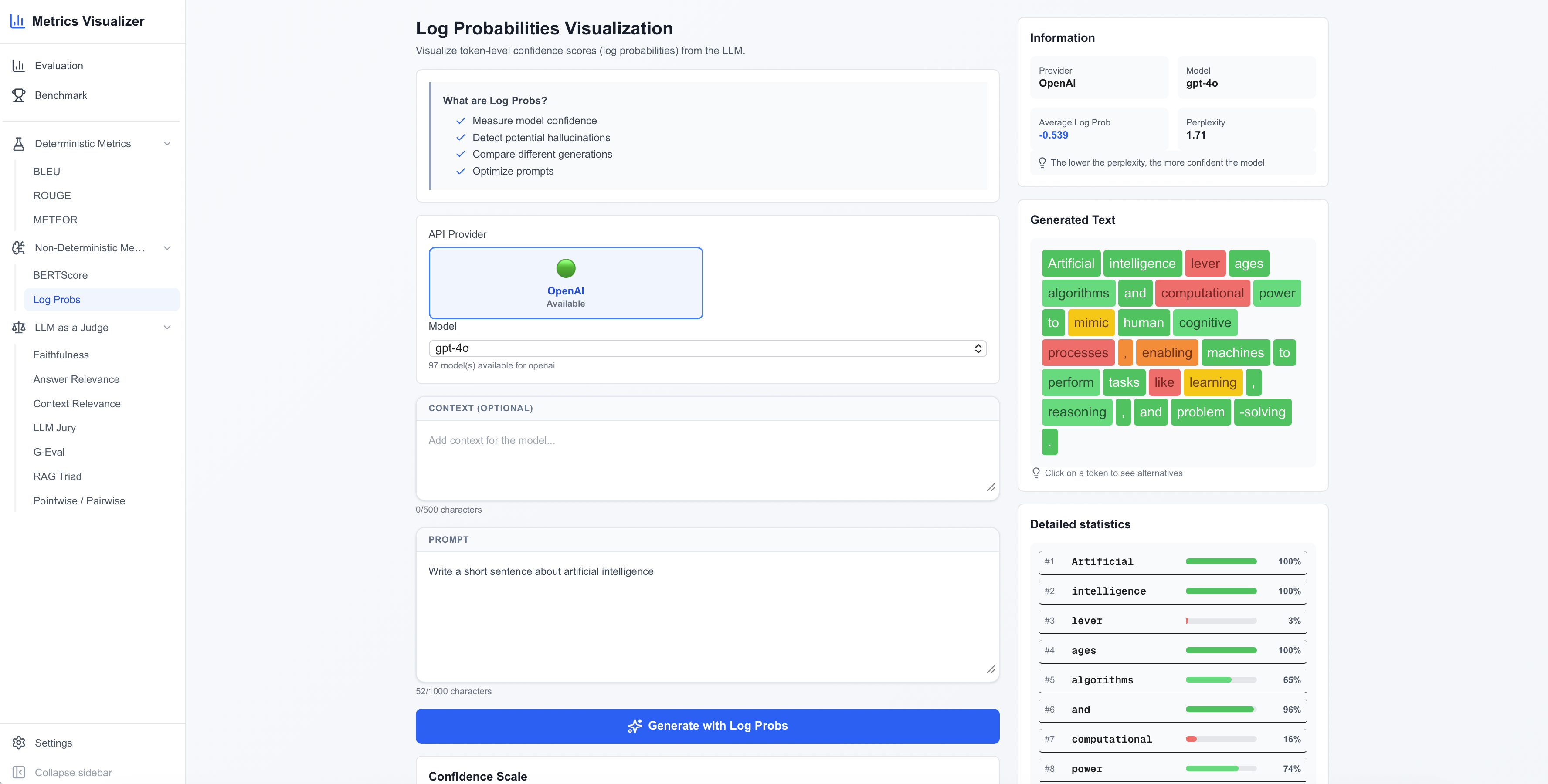}
  \caption{Token-level log-probability visualization: a generated response with per-token confidence color coding (green: high, yellow: moderate, red: low). Low-confidence tokens indicate hallucination-prone regions.}
  \Description{Screenshot of the LogProbs visualization interface. A generated text response is shown with individual tokens highlighted in colors ranging from green to red based on their log-probability scores. A legend explains the color scale.}
  \label{fig:logprobs}
\end{figure}

A machine learning engineer (acting as a Technical Expert) is developing a RAG pipeline for a legal document summarization system. They suspect that the retriever is returning partially irrelevant context and that the generator occasionally introduces unsupported claims. Using LLM-FACETS, the engineer:
\begin{enumerate}
  \item \textbf{Setup inputs}: They submit a question, retrieved context passage, and generated answer to the RAG Triad interface.
  \item \textbf{Observe scores}: They note that the Faithfulness score is 0.6 and Context Relevance is 0.4. All scores are normalized to $[0, 1]$, where 1 indicates the best result (Section~\ref{sec:architecture}): a Faithfulness of 0.6 means 3 of 5 generated claims are unsupported by the context, while a Context Relevance of 0.4 indicates that the retrieval is noisy.
  \item \textbf{Analyze breakdowns}: They use the per-claim breakdown (visible in Figure~\ref{fig:ragtriad}) to identify which specific generated sentence introduces the hallucination.
  \item \textbf{Cross-reference metrics}: They correlate the finding with the LogProbs visualization, noticing the hallucinated phrase corresponds to low-confidence tokens ($p < 0.4$, orange tier), confirming model uncertainty (Figure~\ref{fig:logprobs}).
  \item \textbf{Iterate}: They iterate on the retrieval strategy and re-evaluate without leaving the browser, repeating until Faithfulness exceeds the target threshold (e.g., 0.85) or until unsupported claims are eliminated.
\end{enumerate}
This workflow compresses a debugging loop that previously required scripting, library installation, and manual log analysis into a single browser session, reducing iteration time from hours to minutes.
\subsection{Example 2: The Compliance Officer -- Auditing Without Coding}
A legal compliance officer (acting as a Compliance Officer) at a healthcare organization needs to certify that an LLM-based patient triage assistant does not fabricate medical information. The officer has no programming background. Following the workflow presented in Section~\ref{sec:workflow}:
\begin{enumerate}
  \item \textbf{Define transparency goal}: Factual grounding (are recommendations supported by clinical protocol?).
  \item \textbf{Select metrics}: Faithfulness and Context Relevance (RAG Triad), targeting the factual dimension.
  \item \textbf{Load dataset}: The officer uploads a local CSV of patient queries and model responses. To preserve privacy, they leverage the tool's zero-transmission architecture~(C3), which ensures data query operations are executed entirely in-process via DuckDB while the API key remains stored strictly in the browser's IndexedDB.
  \item \textbf{Cross-check results}: The Faithfulness score (Figure~\ref{fig:ragtriad}) reveals which triage recommendations are unsupported by the retrieved clinical protocol; the per-claim breakdown identifies the specific sentences introducing unsupported claims.
  \item \textbf{Package evidence}: The officer exports a structured JSON artifact containing per-claim scores and reasoning chains for inclusion in the regulatory compliance report.
\end{enumerate}
Unlike the developer's iterative debugging loop (Example~1), the compliance officer's workflow is a single-pass audit: the goal is to produce a complete evidence artifact certifying whether the system meets the required Faithfulness threshold, not to iteratively improve the system. If the audit reveals unacceptable hallucination rates, the officer documents the finding and refers the system back to the development team for remediation.
In this scenario the compliance officer performs an independent, structured evaluation of the AI system without reliance on the development team, using the same methodology but without requiring engineering assistance. This is the operational meaning of responsible AI governance: accountability that does not depend on the goodwill of the system's developers.
\subsection{Example 3: The Domain Expert -- Validating Specialized Knowledge}
A clinical pharmacologist (acting as a Domain Expert) needs to evaluate whether a new open-weight LLM can accurately answer complex drug-interaction questions. The pharmacologist has deep medical expertise but no coding skills. After navigating to the LLM-FACETS instance in their browser (e.g., \texttt{http://localhost:3000}), the expert:
\begin{enumerate}
  \item \textbf{Setup inputs}: They load a tailored benchmark dataset of clinical questions and reference answers into the dataset explorer (Figure~\ref{fig:dataset}) via a simple file upload.
  \item \textbf{Select metrics}: They choose G-Eval (with a custom rubric emphasizing medical accuracy) and Jury to mitigate the evaluation biases of a single model.
  \item \textbf{Analyze outputs}: They identify specific responses where the model diverges from the reference answers by sorting the evaluation results by their G-Eval scores in the benchmark analysis dashboard (Figure~\ref{fig:analysis}).
  \item \textbf{Cross-check divergence}: They check the Jury consensus (Figure~\ref{fig:jury}). If judges strongly disagree on a score, the pharmacologist can manually review the reasoning chains to see if the model's answer is a valid alternative or a hallucinated contradiction.
  \item \textbf{Communicate findings}: They compile the identified failure cases into a structured report for the technical team, documenting the specific clinical edge cases where the LLM fails. This step concludes the evaluation: unlike the developer's iterative workflow, the domain expert's output is a definitive assessment shared with the engineering team for action.
\end{enumerate}
Taken together, these cases illustrate that transparency is not a fixed property of an evaluation tool, but a relation that must be shaped to the specific epistemic needs and accountability roles of each practitioner. The developer requires low-level diagnostic precision; the compliance officer requires evidence portable into a regulatory report; the domain expert requires the ability to interrogate the semantic validity of the model. The same framework serves all three because its mechanisms are oriented toward distinct purposes rather than toward a single technical artifact.
This workflow makes the framework's methodological contribution explicit: it is not sufficient to provide tools; the framework must guide practitioners in applying those tools purposefully to produce evidence that is both scientifically credible and practically actionable.
\section{Discussion}
\label{sec:discussion}
In this section, we reflect on the implications of LLM-FACETS for the broader LLM evaluation ecosystem. We discuss how its extensibility fosters community-driven infrastructure, clarify its stance on data flow transparency, and connect its methodological approach to the goals of open science and responsible computing.
\subsection{Extensibility as Community Infrastructure}
LLM-FACETS is not an evaluation tool with a fixed metric set: it is a plugin infrastructure on which the research community can build. The Metric Registry and Dataset Registry (Section~\ref{sec:architecture}) mean that any metric proposed in a future paper can be integrated into LLM-FACETS by implementing a single interface, without modifying the evaluation pipeline or the dashboards that consume it. This architectural choice directly addresses the onboarding barrier~(C1) and the reproducibility challenge~(C4): it prevents every team from reimplementing the same evaluation scaffolding from scratch, and it enables cumulative progress---a new faithfulness metric published in 2027 can slot into the same workflow, visualization, and export format as the metrics described in this paper.
This extensibility is also what makes the framework genuinely useful for long-term governance contexts. AI regulation evolves: new transparency obligations will create new metric requirements. A fixed tool becomes obsolete; a plugin architecture grows with the field. The same property applies to data: as new benchmarks for bias, fairness, or multimodal evaluation emerge, the Dataset Registry provides an integration path without architectural overhaul.
We view the plugin architecture as the primary technical contribution of this work, distinct from the specific metrics currently implemented. The current metric suite represents one concrete realization of the framework---a specific set of metrics, datasets, and providers assembled to address the transparency challenges identified in Section~\ref{sec:challenges}. The framework itself, however, provides the foundational components (Metric Registry, Dataset Registry, Provider Factory, workflow engine) into which future metric suites can be integrated---for instance, fairness auditing or multimodal evaluation---without modifying the evaluation pipeline, dashboards, or export format.
\subsection{Data Flow Transparency and the BYOK Distinction}
LLM-FACETS does not claim to eliminate data exposure from all LLM evaluation---doing so would be architecturally impossible for any tool that relies on proprietary external LLM-as-a-Judge APIs. However, it addresses the compliance challenge~(C3) by completely structurally separating local computation from external API calls. While other programmatic tools like Ragas also support Bring Your Own Key (BYOK) approaches, they operate within opaque data-processing environments or rely on Python scripts that can easily blend local and remote processing without explicit boundaries. In contrast, LLM-FACETS enforces data sovereignty by design: deterministic metrics (BLEU, ROUGE, METEOR, BERTScore) run entirely in-process on the self-hosted Next.js server. No evaluation data leaves the practitioner's infrastructure for these tasks. For LLM-judge metrics (Jury, RAG Triad, G-Eval, LogProbs), practitioners supply their own API keys via the browser's IndexedDB. The tool acts exclusively as a stateless proxy; it does not log, store, or intercept the payload en route to the provider. This guarantees that practitioners remain the sole data controllers, assuming full responsibility for their proprietary data interactions without an intermediary platform capturing their evaluation traces.
\subsection{Future Outlook and Open Science as Strategy}
As the LLM evaluation toolscape matures, we anticipate a convergence between technical evaluation capabilities and automated compliance reporting. LLM-FACETS is designed to anticipate this shift by positioning multi-stakeholder transparency~(C5) not as a downstream reporting task, but as an integral part of the evaluation lifecycle. The need for tools that decouple AI accountability from the development teams engineering those systems---addressing the accessibility challenge~(C2)---will only grow as AI regulation enforcement deepens~\cite{euaiact2024}.
Open science remains a foundational strategy for this future. Reproducibility~(C4) ensures that published results---whether in scientific literature or regulatory audits---can be independently verified. By open-sourcing the exact methods for evaluating factual grounding, uncertainty, and model bias, LLM-FACETS offers a transparent alternative to proprietary evaluation APIs whose criteria often remain opaque. The codebase and benchmark datasets will be publicly available, inviting community scrutiny and enabling any team with the data (and the API key) to replicate the tests within their own infrastructure.
\subsection{Limitations}
Several limitations warrant acknowledgment. These concern the scope of the metric definitions, the maturity of the tool's evaluation, and potential threats to the validity of the claims made in this paper. They are organized by their context:
\textbf{(Metric definition) LLM-as-a-Judge biases.} Despite the Jury module's bias mitigation, LLM judges remain susceptible to systematic biases correlated with their training data and instruction tuning. No amount of averaging across judges eliminates shared biases common to models of the same generation or provider.
\textbf{(Tool) LogProb calibration.} Log-probability scores are not universally calibrated across providers or quantization levels. A low-probability token in GPT-4o does not mean the same thing as a low-probability token in a smaller open-weight model. Cross-model comparisons of LogProb-based confidence should be made cautiously.
\textbf{(Process) RAG Triad semantic drift.} The Faithfulness and Answer Relevance metrics depend on LLM-generated intermediate outputs (claim decomposition, reverse question generation). Errors in these intermediate steps propagate to the final scores. The tool mitigates this by displaying intermediate reasoning chains, but does not eliminate the risk.
\textbf{(Metric definition) Metric coverage.} The current suite does not include bias detection metrics, fairness metrics, toxicity assessment, or multimodal evaluation. These represent important dimensions of responsible AI that LLM-FACETS does not yet address.
\textbf{(Validation) Absence of formal stakeholder validation.} The practitioner-centered design choices presented in this paper are grounded in prior HCI research~\cite{liao2020questioning,ehsan2021expanding} and illustrated through representative use cases, but have not yet been validated through formal user studies with domain experts or compliance officers. Empirical evaluation of whether LLM-FACETS genuinely enables independent oversight in practice remains an important direction for future work.
\textbf{(Tool) LogProb availability.} The log-probability visualization feature depends on commercial providers that expose per-token log-probabilities in their API responses. Open-weight models served locally via inference frameworks such as Ollama are not yet integrated, limiting the feature to providers that support it natively. Future work will address this gap through local model integration.
\textbf{(Tool) Limited performance evaluation.} The computational performance analysis presented in Section~\ref{sec:performance} covers only deterministic metrics on a single hardware configuration. Comprehensive benchmarking of LLM-as-a-Judge metrics over massive datasets, as well as scalability analysis across larger distributed clusters, remains for future work.
\section{Conclusion}
\label{sec:conclusion}
Engineering transparency in artificial intelligence cannot remain an exclusive exercise for technical experts and software developers. As AI systems permeate society, their evaluation must become a collaborative, multi-stakeholder endeavor. In this paper, we presented LLM-FACETS, an open-source framework that lowers the barrier to the auditing of LLMs.
By abstracting the complexities of diverse evaluation paradigms---from traditional n-gram metrics to RAG-specific hallucination detection and probabilistic confidence visualization---into an accessible web interface, the framework aims to enable domain experts and compliance officers to actively participate in AI validation alongside technical teams.
LLM-FACETS' plugin architecture ensures that the framework can grow with the field: any metric proposed in future research can be integrated by implementing a single TypeScript interface, without modifying the evaluation pipeline, dashboards, or export format. Its architecture makes data flows explicit---deterministic metrics run within the self-hosted server; LLM-judge metrics contact external APIs transparently, with users retaining full control. Validation confirms that this accessibility does not come at the expense of scientific rigor: deterministic implementations maintain numerical parity with canonical Python reference implementations across 18~metric variants (Section~\ref{sec:verification}).
Future work will focus on: (1)~expanding alignment with the EU AI Act transparency requirements, including automated compliance reporting templates; (2)~incorporating explainable AI (XAI) techniques to provide deeper insights into LLM-as-a-Judge decision processes; (3)~integrating local model inference for fully offline evaluation; (4)~developing shared evaluation campaign infrastructure for coordinated multi-institutional auditing; and (5)~building a community metric registry enabling researchers to publish and share metric implementations as reusable plugins.
Ultimately, true transparency in AI requires both transparent models and transparent, accessible methods of evaluation. LLM-FACETS represents a practical step toward building that collaborative evaluation ecosystem.
\begin{acks}
The authors thank the Scriptor Artis team for infrastructure support during development. This work was conducted as part of a research collaboration between Scriptor Artis SAS and the Luxembourg Institute of Science and Technology (LIST), funded by Scriptor Artis.
\end{acks}
\section*{Competing Interests}
Tom Lucas and Barbara Delacroix are affiliated with Scriptor Artis SAS, which develops and maintains the Devana platform (devana.ai); Devana is integrated as an optional LLM provider in LLM-FACETS. The evaluation framework and its methodology are independent of any commercial product. The other authors declare no competing financial interests or personal relationships that could have influenced this work.
\section*{Funding}
This research was funded by Scriptor Artis SAS (France) through a research collaboration agreement with the Luxembourg Institute of Science and Technology (LIST).
\section*{CRediT Authorship Contribution Statement}
\textbf{Tom Lucas:} Conceptualization, Methodology, Software, Validation, Writing -- original draft.
\textbf{Alessio Buscemi:} Methodology, Validation, Writing -- review \& editing.
\textbf{Alfredo Capozucca:} Methodology, Writing -- review \& editing, Supervision.
\textbf{German Castignani:} Project administration, Writing -- review \& editing, Supervision.
\textbf{Barbara Delacroix:} Supervision.
\section*{Declaration of Generative AI Use}
During the preparation of this work the authors used GitHub Copilot and Google Gemini to assist with code development and documentation. After using these tools, the authors reviewed and edited the content as needed and take full responsibility for the content of the published article.
\section*{Data Availability}
The source code and datasets have been anonymized for peer review and are provided as supplementary material. Upon acceptance, the repository will be made publicly available under the MIT License with Commons Clause.
\bibliographystyle{ACM-Reference-Format}
\bibliography{references}
\end{document}